\documentclass[letterpaper]{article} 
\usepackage{aaai2026}  
\usepackage{times}  
\usepackage{helvet}  
\usepackage{courier}  
\usepackage[hyphens]{url}  
\usepackage{graphicx} 
\usepackage{natbib}  
\usepackage{caption} 
\usepackage{algorithm}
\usepackage{algorithmic}

\usepackage{newfloat}
\usepackage{listings}
\DeclareCaptionStyle{ruled}{labelfont=normalfont,labelsep=colon,strut=off} 
\floatstyle{ruled}
\newfloat{listing}{tb}{lst}{}
\floatname{listing}{Listing}

\newcommand{\net}{SparseSurf}
\usepackage{cuted}
\usepackage{bm}
\usepackage{multicol}
\usepackage{multirow}
\usepackage{amsfonts} 
\usepackage{amsmath}
\usepackage{adjustbox}
\usepackage{booktabs}
\usepackage{xcolor,colortbl}
\definecolor{yellow}{rgb}{1, 1, 0.7}
\definecolor{orange}{rgb}{1, 0.85, 0.7}
\definecolor{red}{rgb}{1, 0.7, 0.7}
\definecolor{normalred}{rgb}{1, 0, 0}
\usepackage{pifont}
\newcommand{\cmark}{\ding{51}}%
\newcommand{\xmark}{\ding{55}}%
\definecolor{Dred}{rgb}{0.6,0,0}
\definecolor{Dgreen}{rgb}{0,0.6,0}
\title{SparseSurf: Sparse-View 3D Gaussian Splatting for Surface Reconstruction}
\author {
    Meiying Gu\textsuperscript{\rm 1},
    Jiawei Zhang\textsuperscript{\rm 1},
    Jiahe Li\textsuperscript{\rm 1},
    Xiaohan Yu\textsuperscript{\rm 2},
    Haonan Luo\textsuperscript{\rm 3},
    Jin Zheng\textsuperscript{\rm 1, \rm 4\thanks{Corresponding authors.}},
    Xiao Bai\textsuperscript{\rm 1$^*$}
}
\affiliations {
    \textsuperscript{\rm 1}School of Computer Science and Engineering, State Key Laboratory of Complex Critical Software Environment, Jiangxi Research Institute, Beihang University, China\\
    \textsuperscript{\rm 2}Macquarie University, Australia\\
    \textsuperscript{\rm 3}School of Computing and Artificial Intelligence, Southwest Jiaotong University, China\\
    \textsuperscript{\rm 4}State Key Laboratory of Virtual Reality Technology and Systems, Beijing, China\\
    \{gumeiying, byzhangjw, lijiahe, jinzheng, baixiao\}@buaa.edu.cn, xiaohan.yu@mq.edu.au, lhn@swjtu.edu.cn
}

\begin{document}

\maketitle

\begin{abstract}
Recent advances in optimizing Gaussian Splatting for scene geometry have enabled efficient reconstruction of detailed surfaces from images. However, when input views are sparse, such optimization is prone to overfitting, leading to suboptimal reconstruction quality. Existing approaches address this challenge by employing flattened Gaussian primitives to better fit surface geometry, combined with depth regularization to alleviate geometric ambiguities under limited viewpoints. Nevertheless, the increased anisotropy inherent in flattened Gaussians exacerbates overfitting in sparse-view scenarios, hindering accurate surface fitting and degrading novel view synthesis performance. In this paper, we propose \net{}, a method that reconstructs more accurate and detailed surfaces while preserving high-quality novel view rendering. Our key insight is to introduce Stereo Geometry-Texture Alignment, which bridges rendering quality and geometry estimation, thereby jointly enhancing both surface reconstruction and view synthesis. In addition, we present a Pseudo-Feature Enhanced Geometry Consistency that enforces multi-view geometric consistency by incorporating both training and unseen views, effectively mitigating overfitting caused by sparse supervision. Extensive experiments on the DTU, BlendedMVS, and Mip-NeRF360 datasets demonstrate that our method achieves the state-of-the-art performance. 

\end{abstract}

\begin{links}
    \link{Project}{https://miya-oi.github.io/SparseSurf-project}
\end{links}

\section{Introduction}

Surface reconstruction is a long-standing problem, aiming to recover accurate 3D geometry from images. Traditional multi-view stereo (MVS) pipelines \cite{mvsnet, ding2022transmvsnet} establish dense correspondences and fuse depth maps via volumetric methods or Poisson reconstruction. Neural Radiance Field (NeRF) based methods \cite{neus, volsdf,geoneus,li2023neuralangelo} build neural signed distance fields and extract high-quality surfaces from the optimized fields, while they require long optimization, hindering real-world applications. 

Recently, 3D Gaussian Splatting has been investigated for surface reconstruction \cite{2dgs,yu2024gof,pgsr}, efficiently estimating detailed surfaces with a sufficient collection of camera views. However, in real-world applications, the acquired views can be very sparse. Since 3DGS relies heavily on multi-view correspondences for optimization, it suffers from severe overfitting when the number of training views is limited \cite{li2024dngaussian}, leading to degraded reconstruction quality.

\begin{figure}[t]
\centering
  \includegraphics[width=\linewidth]{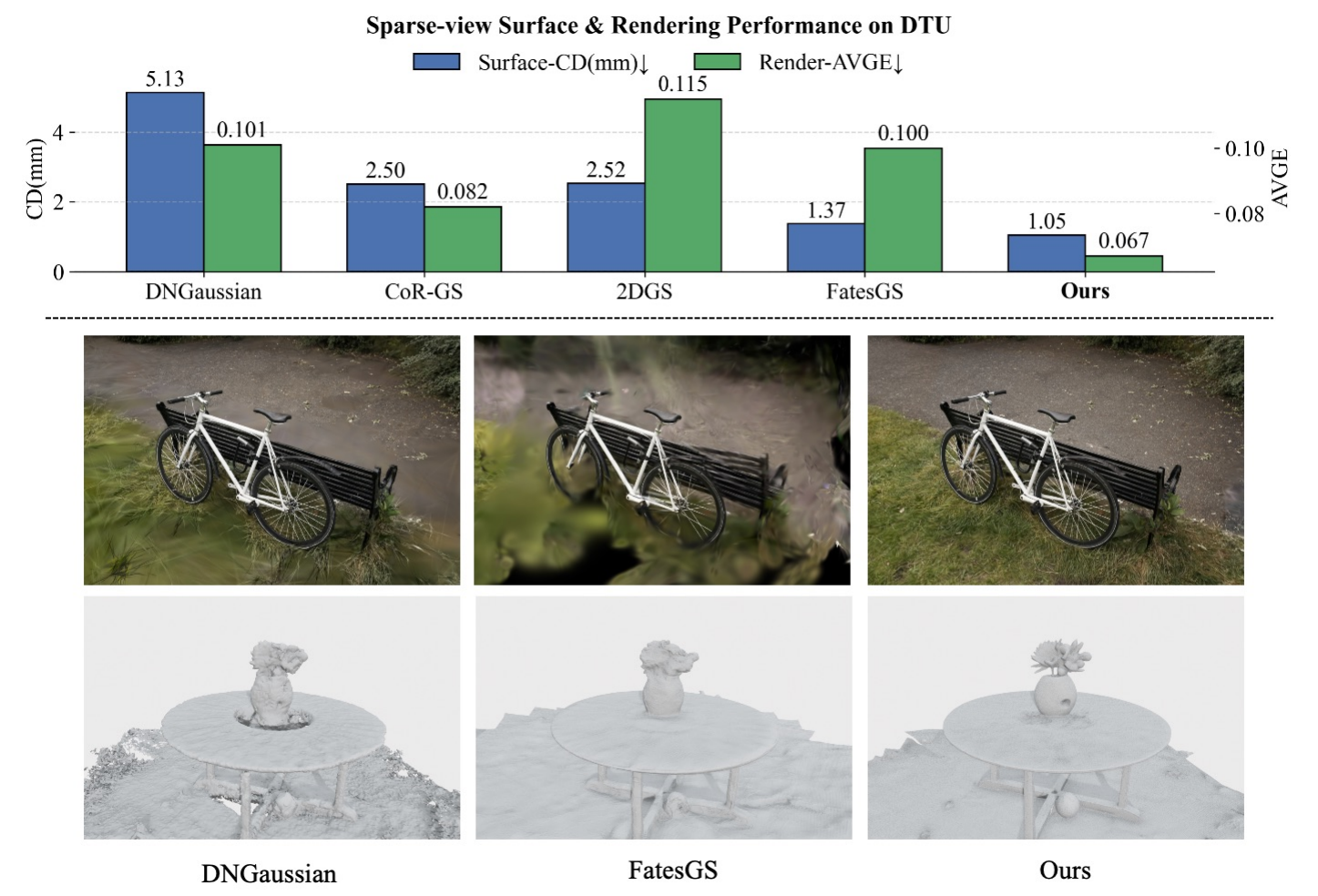}
    \caption{Comparison of sparse-view novel-view synthesis and surface reconstruction on DTU and Mip-NeRF360. Our \net{} achieves the best performance on both surface reconstruction and rendering in sparse-view setting. }
    \label{fig:intro}
\end{figure}

To address the overfitting issues in sparse-view Gaussian Splatting, recent works \cite{zhu2023fsgs,li2024dngaussian,zhang2024cor} have introduced regularization strategies to improve the geometric quality of 3D Gaussians. However, these methods primarily focus on enhancing novel view synthesis (NVS) performance and thus apply relatively loose geometric constraints, making it challenging to reconstruct accurate surfaces. More recently, FatesGS \cite{huang2025fatesgs} and Sparse2DGS \cite{wu2025sparse2dgs} represent Gaussians as flattened 2D planes to better fit surface geometry under sparse views, while enforcing multi-view geometric consistency as regularization. Despite these improvements, the increased anisotropy of flattened Gaussians further amplifies the risk of overfitting in sparse-view settings. As shown in Figure~\ref{fig:intro}, FatesGS with flattened Gaussians meets challenges to render high-fidelity images at unseen views, indicating the flattened Gaussians meet difficulties fitting the accurate
surface positions solely with sparse training view supervision, limiting the detailed surface reconstruction ability.

In this paper, we propose \net{}, which achieves accurate surface reconstruction while preserving photorealistic novel view synthesis. To compensate for the lack of geometry of sparse views, current solutions employ monocular depth priors as additional constraints \cite{li2024dngaussian,huang2025fatesgs}. However, such priors suffer from scale ambiguities and lack confidence estimation, introducing noise into the optimization process. While Gaussians can tolerate noise with sufficient training views, they become significantly susceptible under sparse-view conditions. Prior-induced noise leads to multi-view inconsistency and hinders Gaussians from fitting accurate surfaces. To address this, we introduce Stereo Geometry-Texture Alignment that provides metric supervision, enabling more reliable geometric guidance. These priors are further filtered through geometric consistency to ensure more robust constraints. As training progresses, the quality of stereo renderings improves, which in turn enhances the accuracy of the stereo priors. 

Recovering fine-grained geometry often motivates the use of flattened, anisotropic Gaussian primitives, as this increases surface adherence. However, such flattening reduces the implicit regularization provided by anisotropy, thereby exacerbating overfitting under sparse views in Figure~\ref{fig:intro}. Although overfitting is not readily apparent from the training views, its adverse effects become evident on novel views. To address this challenge, we propose Pseudo-Feature Enhanced Geometry Consistency that enforces multi-view geometric consistency by leveraging both sparse training views and pseudo-unseen views. To recover surface details, we enhance the geometry consistency by introducing multi-view feature representations and conducting feature distillation to efficiently render pseudo view feature representations. By jointly enforcing consistency on training and pseudo-unseen views, our approach mitigates overfitting introduced by limited views and the use of flattened Gaussian primitives, resulting in more robust and detailed surface reconstructions.

Our contributions are summarized as follows.
\begin{itemize}
    \item We propose \net{}, investigating to reconstruct accurate and detailed surfaces while simultaneously improving novel-view synthesis quality of Gaussian Splatting under sparse training views.  
    \item We propose Stereo Geometry-Texture Alignment and Pseudo-Feature Enhanced Geometry Consistency to derive metric depth supervision and enhance pseudo-view through multi-view feature consistency, alleviating overfitting issues and improving the surface reconstruction. 
    \item We perform extensive experiments on DTU, BlendedMVS, and Mip-NeRF360 datasets. Our method achieves state-of-the-art in sparse view surface reconstruction.
\end{itemize}

\section{Related Work}

\subsection{Neural Implicit Representation}
Neural Radiance Fields (NeRF) \cite{nerf} represents scenes as continuous volume radiance functions, achieving remarkably photorealistic rendering. However, the heavy MLPs causes high computational cost.  
Notably, 3D Gaussian Splatting (3DGS) \cite{3dgs} optimizes a set of anisotropic 3D Gaussian primitives with rasterization-based rendering, achieving fast rendering speed and high-quality performance. However, both NeRF and 3DGS focus more on view synthesis than accurate geometry, resulting in messy and noisy surfaces.

\subsection{Neural Surface Reconstruction}
In obtaining a set of 2D images, neural surface reconstruction methods focus on recovering the accurate 3D geometry for the scene. Built upon NeRF, NeuS \cite{neus}, VolSDF \cite{volsdf}, and subsequent optimization methods \cite{monosdf,geoneus,neuralwarp,li2023neuralangelo} represent surfaces as occupancy or signed distance fields. Despite accurate geometry, these implicit surface methods demand significant computational resources and prolonged optimization—the reconstruction time can reach up to several days per scene.

Recent works try to bring the advantages of 3DGS to surface reconstruction. Sugar \cite{guedon2024sugar} introduces regularization for aligning Gaussians with surface, and adopts an optional refinement to bind Gaussians to the mesh surface. 2DGS \cite{2dgs} and Gaussian Surfels \cite{gaussiansurfels} flatten the original 3D gaussians into 2D ellipse to recover thin surfaces. To extract continuous surfaces, GOF \cite{yu2024gof} introduces an opacity field, while GSDF \cite{yu2024gsdf}, 3DGSR \cite{lyu20243dgsr}, and GS-Pull \cite{zhang2024gspull} leverage distance fields. To constrain geometry, PGSR \cite{pgsr} proposes unbiased depth rendering method. Among them, GS2Mesh \cite{wolf2024gs2mesh} is most relevant to our work, as it extracts meshes using a pretrained stereo matching model instead of directly relying on depth maps rendered from the vanilla 3DGS representation. GS2Mesh recovers accurate and complete geometry with dense view inputs. However, under sparse views, the render quality degrades due to overfitting, making it difficult to directly use stereo depth for surface reconstruction.

\begin{figure*}[t]
\centering
  \includegraphics[width=0.95\linewidth]{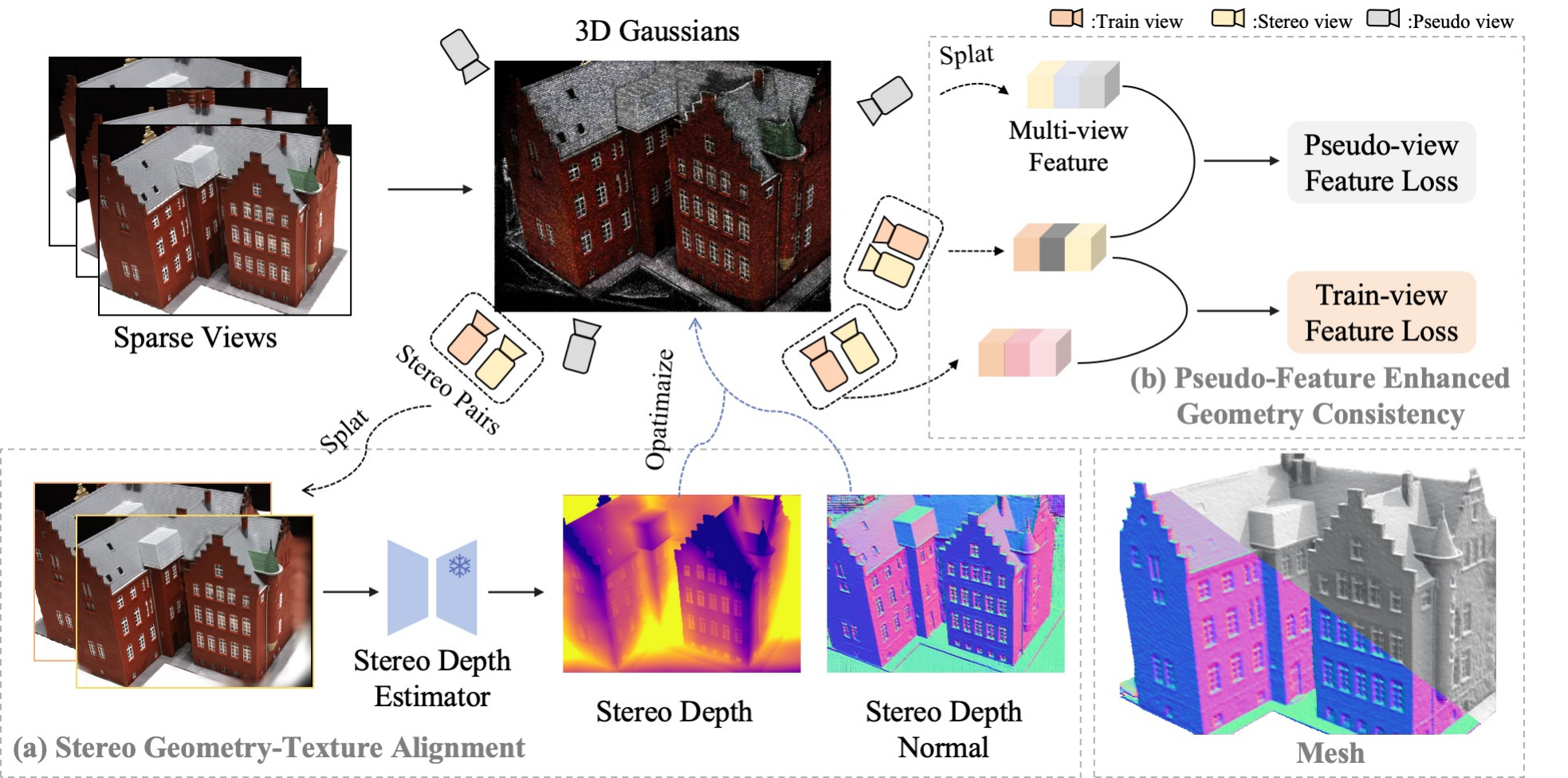}
    \caption{The framework of \net{}. (a) Stereo Geometry-Texture Alignment. We estimate and update stereo-view images to generate binocular priors for geometry supervision. (b) Pseudo-Feature Enhanced Geometry Consistency. To mitigate overfitting and enhance multi-view consistency, we introduce Pseudo-view Feature Consistency and Train-view Feature Alignment. }
    \label{fig:method-overview}
\end{figure*}  
\subsection{Sparse View Gaussian Splatting}
3DGS demonstrates significant advantages in both render quality and inferring speed. While given sparse views, many methods tend to reconstruct noisy and unrealistic scenes. Gaussian splatting faces the problems of overfitting and geometric inaccuracy with limited inputs. Recently, sparse-view novel view synthesis methods aims to alleviate overfitting issue for improve the novel view rendering via geometric priors \cite{li2024dngaussian, zhu2023fsgs, han2024binocular, zheng2025nexusgs} or pseudo views \cite{zhu2023fsgs, zhang2024cor}. 
Although these methods render high-quality appearance of scenes, they are still challenging to recover surfaces. FatesGS \cite{huang2025fatesgs} and concurrent Sparse2DGS \cite{wu2025sparse2dgs} focus on sparse view surface reconstruction. To fit surfaces, they represent 3D scenes as flattened Gaussians under multi-view depth regularization. However, with limited input views, the enhanced anisotropy of flattened Gaussians increases the overfitting risk.

\section{Method}

Given sparse-view RGB images of a scene, our goal is to achieve accurate and detailed surface reconstruction and high-quality rendering. We propose \net{}, as illustrated in Figure \ref{fig:method-overview}.

\subsection{Preliminary of 3D Gaussian Splatting}
\noindent\textbf{Scene Representation.} 
3D Gaussian splatting \cite{3dgs} models a scene as a set of 3D Gaussians. The $i$-th Gaussian is defined by:
\begin{equation}
    \mathcal{G}_i(x) = e^{-\frac{1}{2}(x-\bm\mu_i)^T\bm\Sigma_i^{-1}(x-\bm\mu_i)},
\end{equation}
where $\bm\mu \in \mathbb{R}^3$ , $\bm\Sigma \in \mathbb{R}^{3 \times 3}$, and $x$ presents its center, covariance, 3D position respectively. The covariance matrix $\bm\Sigma$ can be calculated from the scale $\bm{S}$ and rotation $\bm{R}$:
\begin{equation}
\bm{\Sigma}_i = \bm{R}_i\bm{S}_i\bm{S}_i^\top \bm{R}_i^\top
\end{equation}

To accurately conform to the scene surface, we compress the Gaussian ellipsoid along the direction of the minimum scale factor, following \cite{pgsr, gaussiansurfels}. 

\noindent\textbf{Rendering.} For Gaussian rasterization, 3D Gaussians are sorted by depth and renders through alpha-blending. Given a pixel $\bm u$ from one image, the color $\bm{C}\in\mathbb R^3$ can be obtained: 
\begin{equation}
\bm{C} = \sum_{i\in N} T_i \alpha_i \bm{c}_i,\quad T_i=\prod_{j=1}^{i-1}(1 - \alpha_i),
\end{equation}
where $T_i$ is the transmittance, defined as the cumulation of the opacity values of previous Gaussians overlapping the same pixel, and $\alpha_i$ is the blending weight. $\bm{c}_i$ is the view-dependent color.

Similarly, the rendered normal $\bm{N}$ and distance map $\bm{\mathcal{D}}$ can be accumulated as alpha-blending by:
\begin{equation}
    \bm{N}=\sum_{i\in N}\bm{R}_c^T\bm{n}_i\alpha_i \prod_{j=1}^{i-1}(1-\alpha_j),\quad\bm{\mathcal{D}}=\sum_{i\in N}d_i\alpha_i \prod_{j=1}^{i-1}(1-\alpha_j),
\label{eq:distance}
\end{equation}
where $\bm{R}_c$ is the rotation from the camera to the global world, and $\bm{n}_i$ is the minimum scale factor. $d_i=\bm{R}_c^T(\bm{\mu}_i-\bm{T}_c))^T(\bm{R}_c^T\bm{n}_i)$ is the distance from the plane to the camera center, where $\bm{T}_c$ and $\bm{\mu}_i$ are the camera center in the world and the center of gaussian respectively. After obtaining the distance and normal of the plane through rendering, the unbiased depth map $\bm{D}$ can be determined by intersecting rays with the plane \cite{pgsr}.

\subsection{Stereo Geometry-Texture Alignment}

Accurate 3D scene representation requires both reliable rendering and precise geometric reconstruction. Leveraging a binocular framework can combine the advantages of rendering and geometry reconstruction simultaneously. To this end, we render corresponding stereo-view images and feed them into pretrained stereo matching networks \cite{wen2025stereo,zhang2025investigating,stereoanywhere} to obtain precise priors supervising the geometry. As the rendering quality improves, the depth prior becomes more accurate.

\noindent\textbf{Stereo Prior Estimation.} 
3DGS exhibits excellent rendering quality, especially for interpolated viewpoints, yet its performance degrades when rendering extrapolated views. In the case of sparse view, the camera distribution is unbalanced. Therefore, we generate the pseudo-stereo view in the direction of its nearest neighbor view.
Specifically, for each given camera pose $\mathbf{P}_i$, we can obtain a corresponding stereo-view camera at a horizontal baseline $b$. For each train viewpoint, we render its stereo-view image to form a stereo pair, which is then passed through a pre‐trained stereo‐matching network \cite{wen2025stereo} to obtain a disparity map. Using the known camera baseline $b$ and focal length, we convert disparity to depth $\mathcal{D^*}$. To capture local depth variations, we estimate per‐pixel normals $\mathcal{N^*}$ from the depth map $\mathcal{D^*}$ by applying image‐space depth gradients. With the limited input views, the rendered images often exhibit blurring and aliasing that can corrupt geometric priors if used directly. To mitigate this, we apply a stereo view consistency check to produce reliability masks $\mathcal{M^*}$ that filter out unreliable pixels before supervision. Throughout training, at regular intervals, we re-render the stereo-view image, and then update the stereo depth $\mathcal{D^*}$, normal $\mathcal{N^*}$, and valid mask $\mathcal{M^*}$.

\noindent\textbf{Stereo Geometry Supervision.} 
We enforce geometry by minimizing $\mathcal{L}_{1}$ loss between rendered depth ${D}$ and depth prior $\mathcal{D^*}$:
\begin{equation}
\mathcal{L}_{depth} = \mathcal{L}_1({D}, \mathcal{D^*})
\label{eq:stereodepth_loss}
\end{equation}
To sharpen local surface orientation, we employ a cosine‐based loss to align the stereo normal priors $\mathcal{N^*}$ with both the normals ${{N}}$ rendered by 3DGS and those ${N_d}$ computed from the rendered depth.
\begin{equation}
\mathcal{L}_{normal} = 1-\mathcal{C}osine({N}, \mathcal{N^*}), \mathcal{L}_{nd} = 1-\mathcal{C}osine({N}_d, \mathcal{N^*})
\label{eq:sterenormal_loss}
\end{equation}

To transfer the inherent smoothness of our reliable stereo normals onto the rendered normals and normals from depth, we then impose an edge‐aware Laplacian smoothness loss, penalizing deviations between those second‐order gradients and those of the stereo priors.
\begin{equation}
\mathcal{L}_{smooth} = \mathcal{S}mooth({N},\mathcal{N^*})+\mathcal{S}mooth({N}_d,\mathcal{N^*})
\label{eq:smooth_loss}
\end{equation}
The above losses are evaluated only over pixels that pass consistency checks. The total stereo loss can be defined as :
\begin{equation}
\mathcal{L}_{stereo} = (\lambda_d\mathcal{L}_{depth}+\lambda_n\mathcal{L}_{normal}+\lambda_{nd}\mathcal{L}_{nd})\mathcal{M^*}+\lambda_s\mathcal{L}_{smooth}
\label{eq:stereo_loss}
\end{equation}

\subsection{Pseudo-Feature Enhanced Geometry Consistency}

With limited camera views, Gaussian Splatting is prone to overfitting training views when supervised solely by ground‐truth color. By introducing Stereo Geometry-Texture Alignment constrain the arrangement of Gaussian primitives helps to alleviate this effect. However, the high‐quality rendering of vanilla 3DGS relies on anisotropic kernels. To enforce tighter surface adherence, we instead adopt flattened Gaussian kernels, which improve surface quality but sacrifice the innate regularization of anisotropy, bringing extra risk of overfitting in sparse-view supervision. While the overfitting issue are hardly observed from training views, it can be obvious at unseen views due to the high anisotropy of flattened Gaussians. Therefore, we proposed Pseudo-Feature Enhanced Geometry Consistency to regularize the Gaussian geometry, combining both training and unseen views.

\begin{table*}
  \centering
  \label{tab:dtu_little}
  \begin{adjustbox}{width=\textwidth,center}
  \begin{tabular}{l|ccccccccccccccc|c}
        \toprule
        Scan ID & 24 & 37 & 40 & 55 & 63 & 65 & 69 & 83 & 97 & 105 & 106 & 110 & 114 & 118 & 122 & Mean \\
        \midrule
        \midrule
        COLMAP  & 2.88 & 3.47 & \cellcolor{red}1.74 & 2.16 & 2.63 & 3.27 & 2.78 & 3.63 & 3.24 & 3.49 & 2.46 & 1.24 & 1.59 & 2.72 & 1.87 & 2.61 \\
        \midrule
        {SparseNeuS$_{ft}$}  & 4.81 & 5.56 & 5.81 & 2.68 & 3.30 & 3.88 & 2.39 & 2.91 & 3.08 & 2.33 & 2.64 & 3.12 & 1.74 & 3.55 & 2.31 & 3.34 \\ 
        {VolRecon} & 3.05 & 4.45 & 3.36 & 3.09 & 2.78 & 3.68 & 3.01 & 2.87 & 3.07 & 2.55 & 3.07 & 2.77 & 1.59 & 3.44 & 2.51 & 3.02\\
        UFORecon  &  1.52 &  \cellcolor{orange}2.58 &  \cellcolor{yellow}1.85 &  1.44 & 1.55 &  \cellcolor{yellow}1.81 & \cellcolor{yellow}1.06 & 1.52 &  \cellcolor{red}0.96 & 1.40 & \cellcolor{yellow}1.19 & \cellcolor{yellow}0.94 & 0.65 & 1.25 & 1.29 &  1.40 \\
       \midrule
        NeuS  & 4.11 & 5.40 & 5.10 & 3.47 & 2.68 & 2.01 & 4.52 & 8.59 & 5.09 & 9.42 & 2.20 & 4.84 & 0.49 & 2.04 & 4.20 & 4.28  \\
        VolSDF & 4.07 & 4.87 & 3.75 & 2.61 & 5.37 & 4.97 & 6.88 & 3.33 & 5.57 & 2.34 & 3.15 & 5.07 & 1.20 & 5.28 & 5.41 & 4.26 \\
        MonoSDF & 3.47 & 3.61 & 2.10 & 1.05 & 2.37 & \cellcolor{red}1.38 & 1.41 & 1.85 & 1.74 & 1.10 & 1.46 & 2.28 & 1.25 & 1.44 & 1.45 & 1.86 \\
        NeuSurf & \cellcolor{orange}1.35 & 3.25 & 2.50 &  \cellcolor{orange}0.80 & \cellcolor{orange} 1.21 & 2.35 &  \cellcolor{orange}0.77 &  \cellcolor{orange}1.19 &  \cellcolor{yellow}1.20 &  \cellcolor{yellow}1.05 &  \cellcolor{orange}1.05 & 1.21 &  \cellcolor{orange}0.41 &  \cellcolor{orange}0.80 & \cellcolor{yellow} 1.08 &  \cellcolor{orange}1.35 \\
        \midrule
        DNGaussian & 3.40 & 5.58 & 3.28 &4.62&3.18&2.70&6.19&4.84&9.04&3.29&8.10&11.28& 3.90&4.86&2.69&5.13\\
        CoR-GS &2.80 &3.06 &2.42 &1.88&2.37&2.76&2.00&5.66&2.51&1.92&2.54&2.20&1.36&2.28&1.74&2.50\\
        Binocular3DGS &2.13 &3.33 &2.76 &1.33&3.55&2.66&2.48&3.79&2.30&2.74&2.32&5.79&1.53&2.03&1.98&2.71\\
        2DGS  & 3.25 & 3.64 & 3.52 & 1.42 & 2.04 & 2.52 & 1.99 & 2.69 & 2.55 & 1.79 & 2.92 & 4.50 & 0.73 & 2.38 & 1.79 & 2.52 \\
        PGSR &5.83 &4.59 &4.52 &3.36&4.25&3.75&2.81&5.92&4.60&4.27&3.61&6.09&1.02&2.55&2.32&3.97\\
        FatesGS &  \cellcolor{red}1.32 &  \cellcolor{yellow}2.85 &  2.71 &  \cellcolor{orange}0.80 &  \cellcolor{yellow}1.44 & 2.08 &  1.11 &  \cellcolor{orange}1.19 &  1.33 &  \cellcolor{red}0.76 &  1.49 &  \cellcolor{red}0.85 &  \cellcolor{yellow}0.47 & \cellcolor{yellow} 1.05 &  \cellcolor{orange}1.06 & \cellcolor{yellow}1.37 \\
        
        \midrule
        \textbf{Ours} &  \cellcolor{orange}1.35 &  \cellcolor{red}2.10&  \cellcolor{orange}1.84 &  \cellcolor{red}0.66 &  \cellcolor{red}0.79 &  \cellcolor{orange}1.62 &  \cellcolor{red}0.62 &  \cellcolor{red}1.13 & \cellcolor{orange}1.06 &  \cellcolor{red}0.76 &  \cellcolor{red}0.90 &  \cellcolor{orange}0.87 &  \cellcolor{red}0.39 &  \cellcolor{red}0.70 & \cellcolor{red}0.96 &  \cellcolor{red}1.05 \\
        \bottomrule
    \end{tabular}
    \end{adjustbox}
\caption{The quantitative comparisons of CD$\downarrow$ on DTU dataset(little-overlap setting). Best results are highlighted as \colorbox{red}{1st}, \colorbox{orange}{2nd} and \colorbox{yellow}{3rd}.}
\end{table*}

\begin{figure*}[!t]
\centering
  \includegraphics[width=0.95\linewidth]{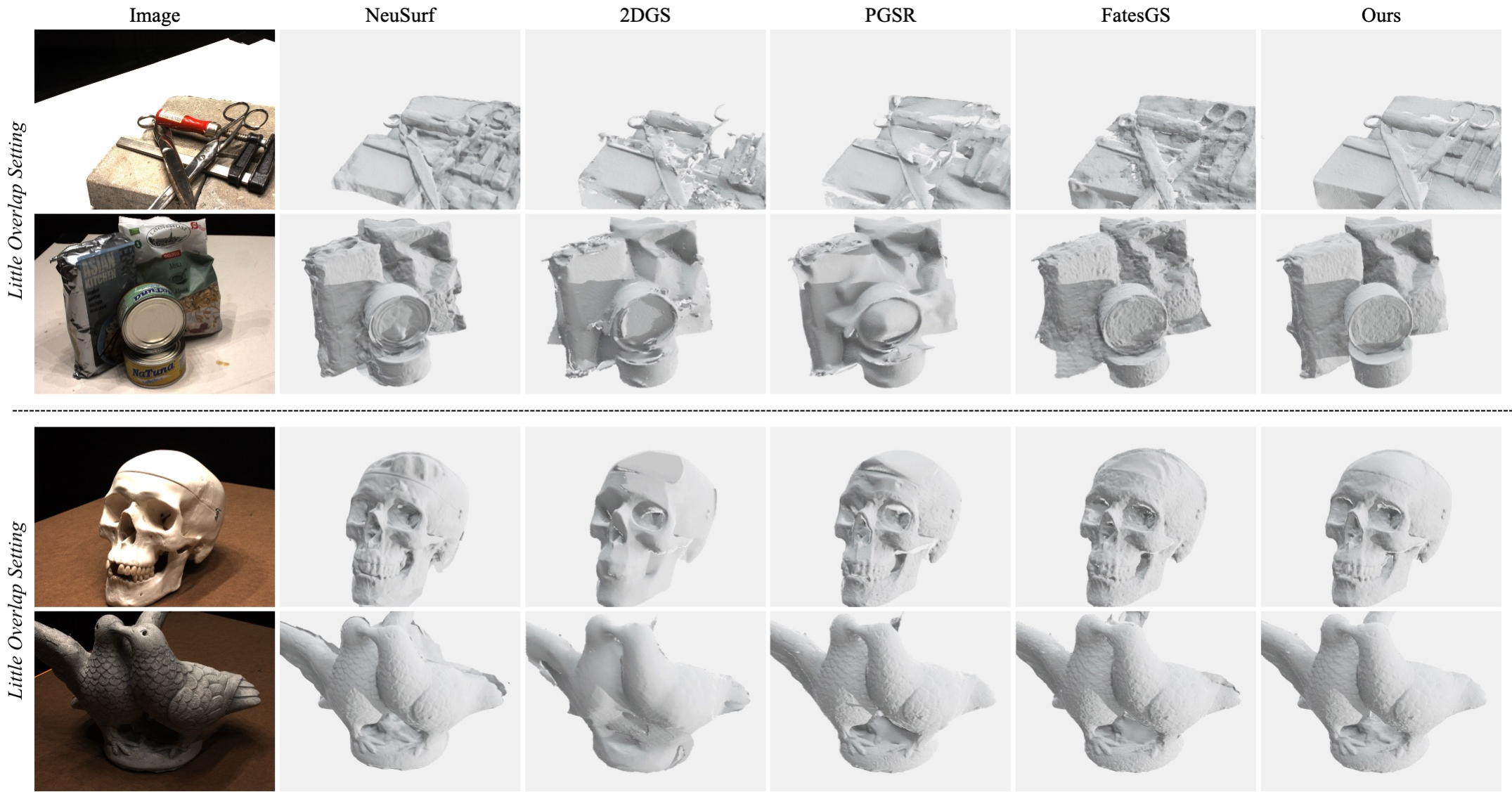}
    \caption{Qualitative comparison of reconstruction results on the DTU with little-overlap and large-overlap sparse setting.}
    \label{fig:dtu_large}
\end{figure*}

\begin{table*}[t]
  \centering
  \begin{adjustbox}{width=\textwidth,center}
  \begin{tabular}{l|ccccccccccccccc|c}
        \toprule
        Scan ID & 24 & 37 & 40 & 55 & 63 & 65 & 69 & 83 & 97 & 105 & 106 & 110 & 114 & 118 & 122 & Mean \\
        \midrule
        \midrule
        COLMAP  & 0.90 & 2.89 & 1.63 & 1.08 & 2.18 & 1.94 & 1.61 & 1.30 & 2.34 & 1.28 & 1.10 & 1.42 & 0.76 & 1.17 & 1.14 & 1.52 \\
        TransMVSNet  & 1.07 & 3.14 & 2.39 & 1.30 & 1.35 & 1.61 & {0.73} & 1.60 & 1.15 & 0.94 & 1.34 & \cellcolor{orange}0.46 & 0.60 & 1.20 & 1.46 & 1.35 \\
        \midrule
        {SparseNeuS$_{ft}$} & 1.29 & {2.27} & 1.57 & 0.88 & 1.61 & 1.86 & 1.06 & 1.27 & 1.42 & 1.07 & 0.99 & 0.87 & 0.54 & 1.15 & 1.18 & 1.27 \\
        {VolRecon}  & 1.20 & 2.59 & 1.56 & 1.08 & 1.43 & 1.92 & 1.11 & 1.48 & 1.42 & 1.05 & 1.19 & 1.38 & 0.74 & 1.23 & 1.27 & 1.38 \\
        ReTR & 1.05 & {2.31} & 1.44 & 0.98 & 1.18 & {1.52} & 0.88 & 1.35 & 1.30 & 0.87 & 1.07 & 0.77 & 0.59 & 1.05 & 1.12 & 1.17 \\
        C2F2NeuS  & 1.12 & 2.42 & {1.40} & \cellcolor{orange}{0.75} & 1.41 & 1.77 & 0.85 & \cellcolor{yellow}{1.16} & 1.26 & 0.76 & 0.91 & 0.60 & {0.46} & 0.88 & {0.92} & 1.11 \\
        GenS$_{ft}$  & 0.91 & 2.33 & 1.46 & \cellcolor{orange}0.75 & \cellcolor{red}1.02 & 1.58 & {0.74} &\cellcolor{yellow} {1.16} & \cellcolor{yellow}{1.05} & 0.77 & {0.88} & {0.56} & 0.49 & 0.78 & 0.93 & 1.03 \\
        UFORecon  & \cellcolor{yellow}0.76 & \cellcolor{yellow}2.05 & \cellcolor{yellow}1.31 & 0.82 & {1.12} & \cellcolor{orange}1.18 & {0.74} & 1.17 & 1.11 & {0.71} & {0.88} & 0.58 & 0.54 & 0.86 & 0.99 & \cellcolor{yellow}0.99 \\
       \midrule
        NeuS  & 4.57 & 4.49 & 3.97 & 4.32 & 4.63 & 1.95 & 4.68 & 3.83 & 4.15 & 2.50 & 1.52 & 6.47 & 1.26 & 5.57 & 6.11 & 4.00  \\
        VolSDF  & 4.03 & 4.21 & 6.12 & 0.91 & 8.24 & 1.73 & 2.74 & 1.82 & 5.14 & 3.09 & 2.08 & 4.81 & 0.60 & 3.51 & 2.18 & 3.41 \\
        MonoSDF & 2.85 & 3.91 & 2.26 & 1.22 & 3.37 & 1.95 & 1.95 & 5.53 & 5.77 & 1.10 & 5.99 & 2.28 & 0.65 & 2.65 & 2.44 & 2.93 \\
        NeuSurf  & {0.78} & 2.35 & 1.55 & \cellcolor{orange}0.75 & \cellcolor{orange}1.04 & 1.68 & \cellcolor{orange}0.60 & \cellcolor{orange}1.14 & \cellcolor{red}0.98 & \cellcolor{yellow}0.70 & \cellcolor{orange}0.74 & {0.49} & \cellcolor{orange}0.39 & \cellcolor{orange}0.75 & \cellcolor{red}0.86 & \cellcolor{yellow}0.99 \\
        \midrule
        3DGS & 3.38 & 4.19 & 2.99 & 1.76 & 3.38 & 3.80 & 5.21 & 2.91 & 4.29 & 3.18 & 3.23 & 5.18 & 2.78 & 3.48 & 3.32 & 3.54 \\
        Gaussian Surfels  & 3.56 & 5.42 & 3.95 & 3.68 & 4.61 & 2.72 & 4.42 & 5.22 & 4.71 & 3.46 & 4.07 & 5.42 & 2.44 & 3.27 & 4.00 & 4.06 \\
        2DGS  & 1.26 & 2.95 & 1.73 & 0.96 & 1.68 & 1.97 & 1.58 & 1.87 & 2.50 & 1.02 & 1.93 & 1.91 & 0.72 & 1.85 & 1.37 & 1.69 \\
        PGSR &1.20 &3.09 &1.97 &1.36&4.99&2.13&1.18&2.43&2.17&1.81&1.36&0.83&0.43&1.37&0.96&1.82\\
        FatesGS  & \cellcolor{red}0.67 & \cellcolor{orange}1.94 & \cellcolor{red}1.17 & 0.77 & 1.28 & \cellcolor{yellow}1.23 & \cellcolor{yellow}0.63 & \cellcolor{red}1.05 & \cellcolor{red}0.98 & \cellcolor{orange}0.69 &\cellcolor{yellow} 0.75 & \cellcolor{yellow}0.48 & \cellcolor{yellow}0.41 & \cellcolor{yellow}0.78 & \cellcolor{yellow}0.90 & \cellcolor{orange}0.92 \\
        Sparse2DGS  &1.05&2.35&1.38&0.83&1.37&1.45&0.84&1.16&1.43&0.74&0.85&0.84&0.57&0.95&1.01&1.13\\
        \midrule
        \textbf{Ours} & \cellcolor{orange}0.72 & \cellcolor{red}1.61& \cellcolor{red}1.17 & \cellcolor{red}0.72 & \cellcolor{yellow}1.11 & \cellcolor{red}1.13 & \cellcolor{red}0.57 & 1.30 & 1.20 & \cellcolor{red}0.67 & \cellcolor{red}0.70 & \cellcolor{red}0.45 & \cellcolor{red}0.36 & \cellcolor{red}0.70 & \cellcolor{orange}0.87 & \cellcolor{red}0.89 \\
        \bottomrule
    \end{tabular}
    \end{adjustbox}
\caption{The quantitative comparisons of CD$\downarrow$ on DTU dataset(large-overlap setting). Best results are highlighted as \colorbox{red}{1st}, \colorbox{orange}{2nd} and \colorbox{yellow}{3rd}.}
\label{tab:dtu_large}
\end{table*}

\begin{figure}[t]
\centering
  \includegraphics[width=1.0\linewidth]{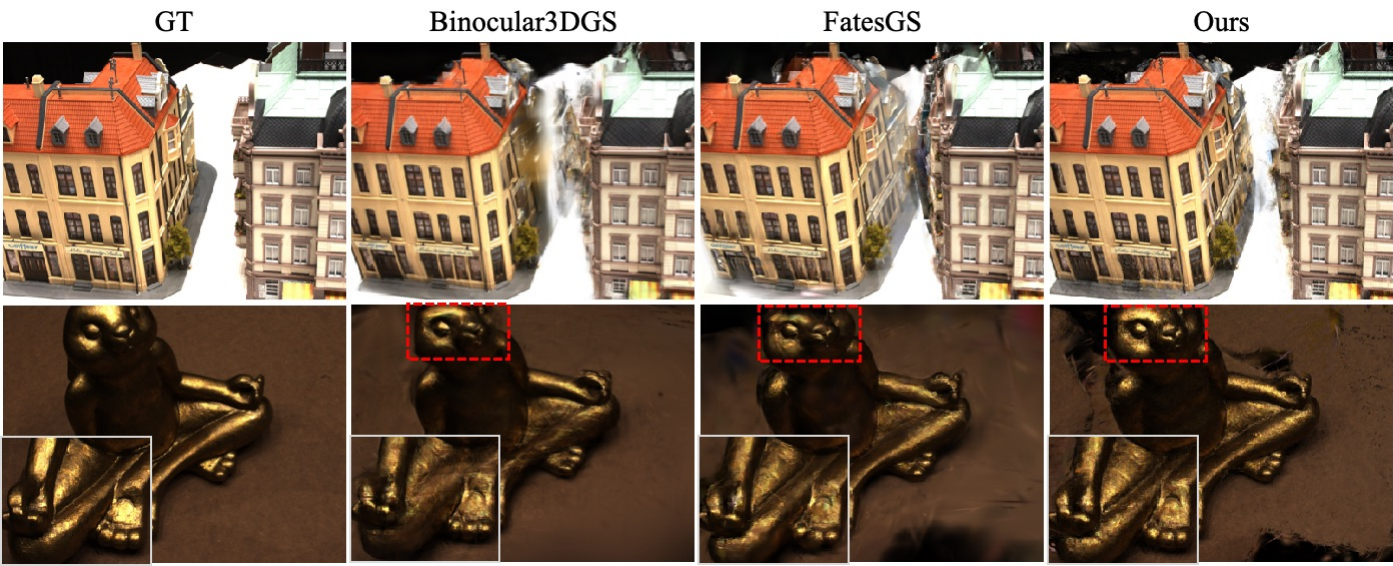}
    \caption{Qualitative rendering comparison on DTU with sparse-view NVS setting.} 
    \label{fig:dtu_nvs}
\end{figure}

\begin{figure}
    \centering
    \includegraphics[width=1.0\linewidth]{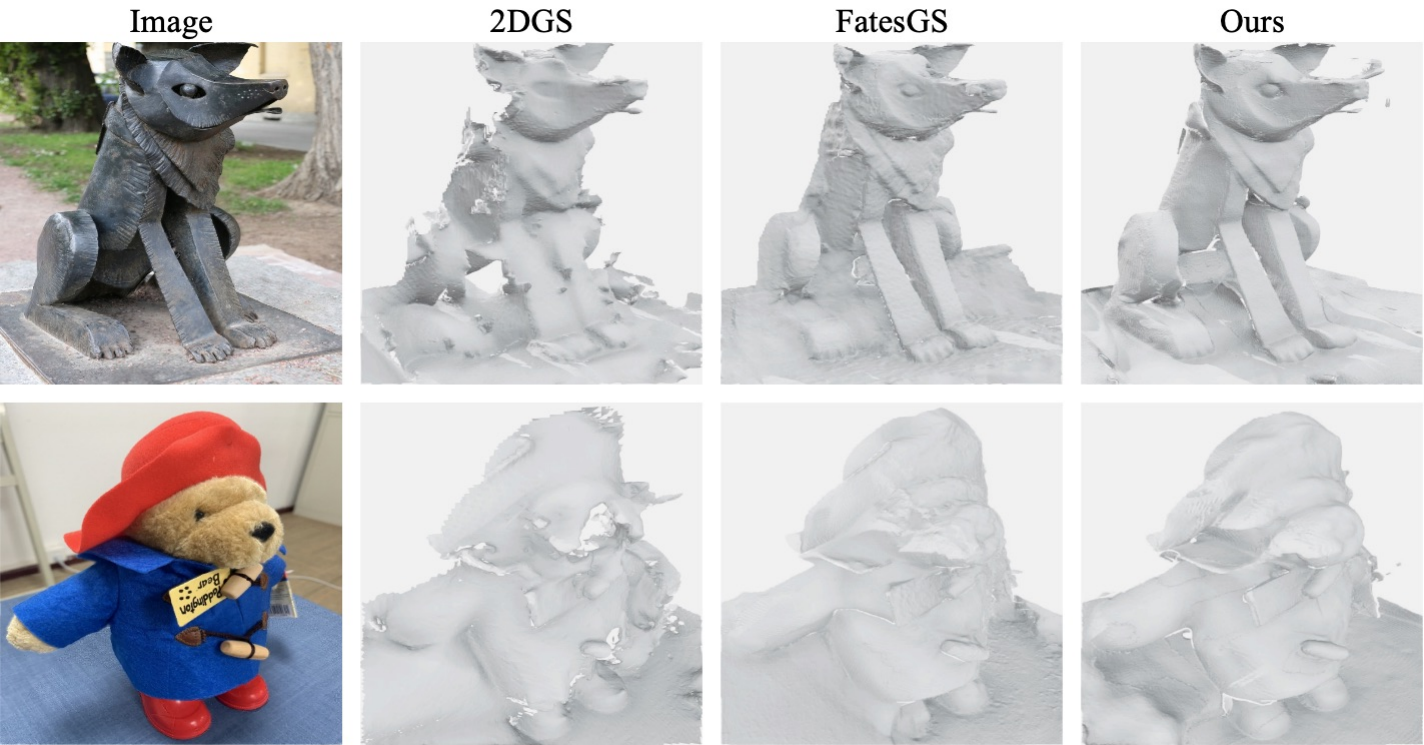}
    \captionof{figure}{Comparison on BlendedMVS.}
    \label{fig:blendedmvs_visual}
\end{figure}

\noindent\textbf{Pseudo-view Feature Consistency.} In sparse-view settings, pseudo-view supervision is regarded as an effective solution to alleviate overfitting because the virtual view enriches the limited set of observed views. Previous NVS works generate pseudo views based on the positions of training cameras and supervise pseudo views with RGB colors \cite{zhang2024cor} or monocular depth maps \cite{zhu2023fsgs}. However, they are insufficient to regularize the surface geometry of pseudo views due to the lack of multi-view constraints. To this end, we supervise pseudo views with multi-view geometry consistency and enhance the regularization effects on details through multi‐view feature representations. While naively extracting features from unseen views at each training loop incurs prohibitive computational cost, we augment each Gaussian primitive with an additional feature attribute to efficiently constrain the feature consistency on unseen views.

Specifically, given a ground-truth image $\hat{I}$, we can extract the multi-view feature $\mathcal{F^*}$ using a frozen feature extraction model. During training, we distill a predicted feature map ${F}$ by minimizing a feature-distillation loss:

\begin{equation}
\mathcal{L}_{f} = 1-\mathcal{C}osine({F}, \mathcal{F^*})
\end{equation}

In this way, the Gaussian primitives learn to distill and reproduce rich multi-view cues directly from the extracted features, and we enforce pseudo-view supervision in a learned feature space.

Specifically, we render feature map ${F_p}$ under a random pseudo-view $\mathcal{V}_p$. Given a reference training view $\mathcal{V}_t$ with corresponding render depth ${D}_t$ and camera pose $\mathcal{P}_t$, the warped feature map $\mathcal{F}_{p2t}$ from $\mathcal{V}_p$ to $\mathcal{V}_t$ can be defined as:

\begin{equation}
\mathcal{F}_{p2t} = \mathcal{W}arp({F}_p, {D}_t, \mathcal{P}_t, \mathcal{P}_p)
\end{equation}

Similarly, we warp the reference feature map $\mathcal{F}_{r}$
into the pseudo‐view to obtain $\mathcal{F}_{t2p}$. By comparing these bidirectional warps via a round‐trip feature discrepancy, we derive a binary confidence mask $\mathcal{M}_{feat}$ that marks only those pixels whose feature reprojection error falls below a threshold.

However, because pseudo‐views can span wide baselines, the rendered feature maps may contain low‐fidelity regions, if enforced at the pixel level, could corrupt the train‐view features. To guard against this, we impose our feature‐consistency loss at the patch level. Concretely, we tile both the warped feature map $\mathcal{F}_{p2t}$
and the reference map $\mathcal{F}_{t2p}$ into local patches, aggregate each patch into a single descriptor via average pooling, and compute a patch‐wise cosine similarity. We then weight each patch loss by its geometric confidence $M_{feat}$ before summing:
\begin{equation}
\mathcal{L}_{pseudo} = \sum_{i,j}\mathcal{M}_{feat}^{(i,j)}[1-\mathcal{C}osine(\bar{\mathcal{F}}^{(i,j)}_{p2t}, \bar{\mathcal{F}}^{(i,j)}_{r}],
\end{equation}
where $\bar{F}^{(i,j)}$ denotes the pooled descriptor of $(i,j)$th patch.

\noindent\textbf{Train-view Feature Alignment.} 
Building on our pseudo-view feature supervision, we further leverage the high-confidence features extracted from ground-truth training views to enforce multi-view consistency at the pixel level. Concretely, let $\mathcal{V}_s$ be one of the source views with feature $\mathcal{F}_s$ and pose $\mathcal{P}_s$. We perform a one-way warp of the source features into the current training frame $\mathcal{V}_t$ to obtain $\mathcal{F}_{s2t}$. We then apply a pixel-wise cosine loss to align the warped and reference features:
\begin{equation}
\mathcal{L}_{train} = 1-\mathcal{C}osine(\mathcal{F}_{s2t}, \mathcal{F}_{s})
\end{equation}

\section{Experiments}

\subsection{Setup}

\noindent\textbf{Datasets.}
We conduct our experiment on three datasets, DTU \cite{jensen2014large}, BlendedMVS \cite{blendedmvs}, and MipNeRF360 \cite{mipnerf360}. The DTU dataset comprises 128 scenes, each with 49 or 69 images. For surface reconstruction, we select 15 scenes to train and evaluate our model on 3 views of both the little-overlap setting (PixelNeRF) and the large-overlap (SparseNeuS) setting, following the previous work \cite{huang2024neusurf, huang2025fatesgs}. We also show the qualitative
reconstruction comparisons on the BlendedMVS and MipNeRF360 datasets. For rendering comparison, we follow the sparse NVS setting on DTU (3 views) and Mip-NeRF360 (24 views) used in previous works \cite{li2024dngaussian, zhu2023fsgs, zhang2024cor, han2024binocular, zheng2025nexusgs}. Consistent with previous sparse-view settings, the camera poses are assumed to be known. 

\noindent\textbf{Evaluation Metrics.}
To evaluate the accuracy of the reconstructed meshes, we use Chamfer Distance (CD) on DTU, using the official evaluation script.
To evaluate the rendering quality, we use PSNR, SSIM, LPIPS, and AVGE scores.

\noindent\textbf{Implementations.}
Following previous research, we use COLMAP \cite{colmap} for Gaussians initialization with additional multi-view features in 8 dimensions. We render stereo-view images and feed them into the advanced stereo matching network \cite{wen2025stereo} to predict geometric priors from 500 iterations and update every 300 iterations. Following previous studies \cite{huang2024neusurf,huang2025fatesgs}, We adopt Vis-MVSNet \cite{vismvsnet} to extract feature representations.

\begin{figure*}[t]
\centering
  \includegraphics[width=0.95\linewidth]{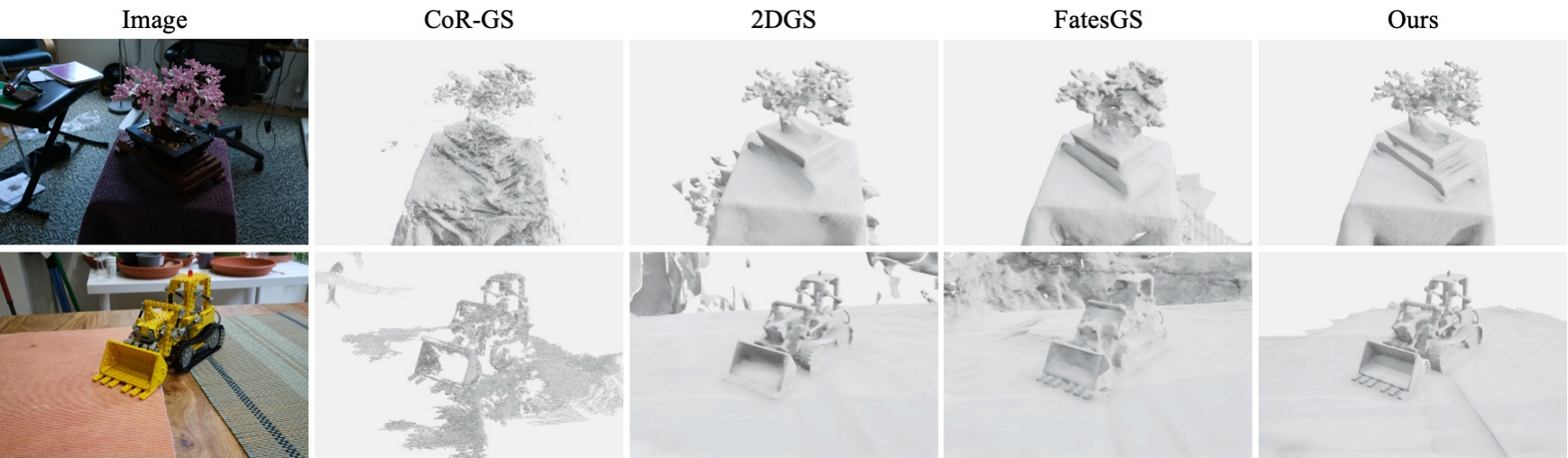}
    \caption{Visual comparison of reconstruction results on Mip-NeRF360 under 24 input views.} 
    \label{fig:m360}
\end{figure*}

\subsection{Comparisons}

\noindent\textbf{Sparse View Surface Reconstruction.}
Quantitative results for geometry reconstruction under sparse input views on both DTU sparse-view settings are reported in Table~\ref{tab:dtu_large}. Our method achieves the best average Chamfer Distance (CD), with a score of 1.05 in the little-overlap setting and 0.89 in the large-overlap setting, outperforming all baselines, including previous state-of-the-art approaches. As shown in Figure~\ref{fig:dtu_large}, our reconstructed meshes exhibit superior geometry in terms of accuracy, completeness, and fine-level detail. Compared with the neural implicit method NeuSurf, our approach produces sharper and more clearly defined edges.

Qualitative results on the BlendedMVS and Mip-NeRF360 datasets are presented in Figure~\ref{fig:blendedmvs_visual} and Figure~\ref{fig:m360}, respectively. Our method demonstrates higher reconstruction accuracy and fidelity on complex real-world scenes.

\begin{table}
    \centering
    \tabcolsep=0.4cm
        \begin{adjustbox}{width=0.5\textwidth,center} 
            \begin{tabular}{@{}l|cccc}
                \toprule
                {Method}     & 
                PSNR$\uparrow$ & SSIM$\uparrow$ & LPIPS$\downarrow$ & AVGE$\downarrow$   \\
                \midrule
                Mip-NeRF            & 8.68  & 0.571 & 0.353 & 0.315 \\
                DietNeRF          & 11.85 & 0.633 & 0.314 & 0.232 \\
                RegNeRF       & 18.89 & 0.745 & 0.190 & 0.107 \\
                FreeNeRF         & 19.92 & 0.787 & 0.182 & 0.095 \\
                SparseNeRF    & 19.55 & 0.769 & 0.201 & 0.102 \\ 
                DS-NeRF             & 16.29 & 0.559 & 0.451 & 0.192 \\       
                ViP-NeRF           & 10.20 & 0.301 & 0.363 & 0.307  \\ 
                SimpleNeRF      & 11.94 & 0.387 & 0.286 & 0.243  \\ 
                \midrule
                3DGS                  & 17.65 & 0.816 & 0.146 & 0.102  \\
                DNGaussian      & 18.91 & 0.790 & 0.176 & 0.101 \\
                FSGS                  & 17.14 & 0.818 & 0.162 & 0.110 \\       
                CoR-GS             & 19.21 & 0.853 & 0.119 & \cellcolor{yellow}0.082 \\     Binocular3DGS              & \cellcolor{orange}20.71 & \cellcolor{yellow}0.862 & \cellcolor{yellow}0.111 & - \\  
                NexusGS  &  \cellcolor{yellow}20.21 & \cellcolor{orange}0.869 & \cellcolor{orange}0.102  & \cellcolor{orange}0.071 \\
                \midrule
                FatesGS    &  17.95 & 0.833 & 0.129  & 0.100 \\
                \textbf{Ours}   &  \cellcolor{red}21.31 & \cellcolor{red}0.886 & \cellcolor{red}0.089  & \cellcolor{red}0.067 \\
                \bottomrule
            \end{tabular}
            \end{adjustbox}
            
        \caption{Quantitative evaluations of render quality on the DTU dataset(Sparse-view novel view synthesis setting). }
        \label{tab:dtu_nvs}    
\end{table}

\noindent\textbf{Sparse View Novel View Synthesis.}
We evaluate our method on DTU with the sparse-view NVS setting for the validation of rendering quality. As shown in Table \ref{tab:dtu_nvs}, compared to the current state-of-the-art sparse-view methods, our approach achieves the best performance across PSNR, SSIM, LPIPS, and AVGE on DTU. Figure \ref{fig:dtu_nvs} illustrates the visual comparison. With the correct geometry and overfit relief strategy, our method has fewer artifacts and less geometric misalignment.

\begin{figure}[t]
\centering
  \includegraphics[width=1.0\linewidth]{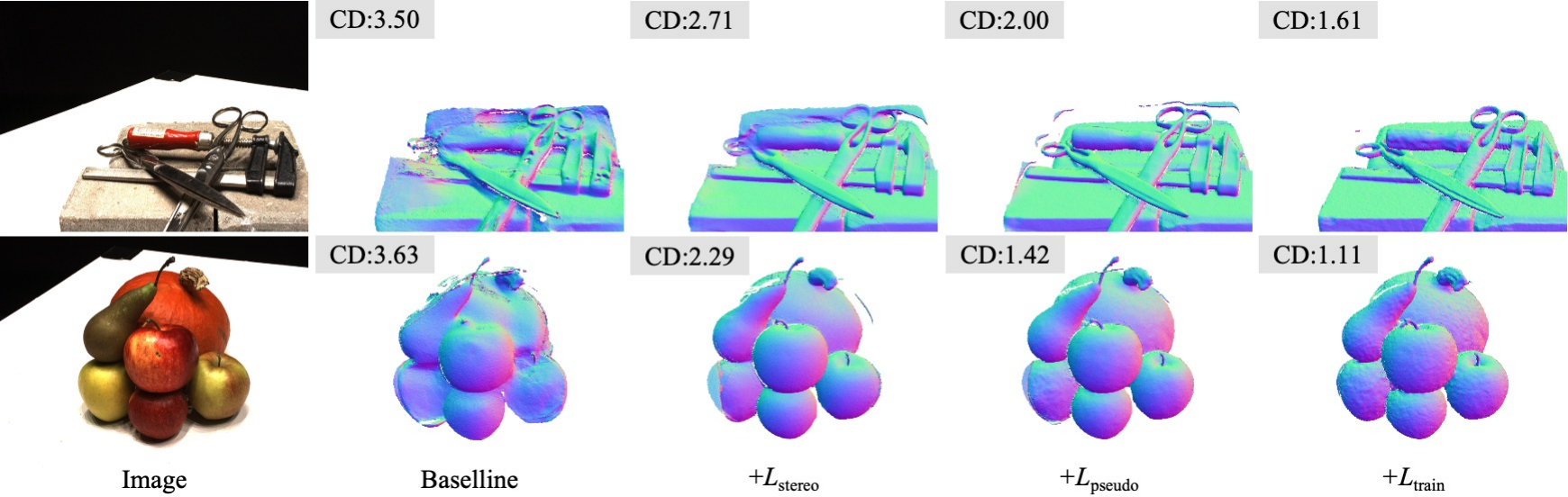}
    \caption{Visualization results for ablating each component of our method.}
    \label{fig:dtu_ablation}
\end{figure}  

\subsection{Ablation Study}

We conduct ablation experiments on DTU dataset with large-overlap setting. To demonstrate the effectiveness of the stereo loss $L_{stereo}$, pseudo-view feature loss $L_{pseudo}$, and train-view feature loss $L_{train}$, we isolate these modules and measure their impact in Table \ref{tab:ablation} and Figure \ref{fig:dtu_ablation}. Additional results and analysis (e.g., efficiency) are presented in the supplementary materials. 

\begin{table}[t] 
  \centering
  \tabcolsep=0.1cm
  \resizebox{1.0\linewidth}{!}{
  \begin{tabular}{ccc|ccc}
    \toprule
       {$L_{stereo}$} & {$L_{pseudo}$} & {$L_{train}$}  &   Accuracy$\downarrow$ & Completion$\downarrow$ & Average$\downarrow$  \\
    \midrule
    \textcolor{Dred}{\xmark}& \textcolor{Dred}{\xmark}&\textcolor{Dred}{\xmark} & 1.318   &2.302& 1.810  \\
    \textcolor{Dgreen}{\cmark} &\textcolor{Dred}{\xmark} & \textcolor{Dred}{\xmark}&0.822 & 1.612 & 1.217 \\
     \textcolor{Dgreen}{\cmark}& \textcolor{Dgreen}{\cmark} & \textcolor{Dred}{\xmark}  & 0.610 & 1.327& 0.969   \\ 
     \textcolor{Dgreen}{\cmark} &  \textcolor{Dgreen}{\cmark}   & \textcolor{Dgreen}{\cmark} & \textbf{0.533} &  \textbf{1.239}   & \textbf{0.886}  \\
  \bottomrule
  \end{tabular}
 }
 \caption{Ablation Study for the large-overlap setting on the DTU dataset.}
\label{tab:ablation}
\end{table}

\noindent\textbf{Stereo Geometry-Texture Alignment.} 
Compared to the baseline method, incorporating binocular depth supervision significantly improves the surface quality from an average distance 1.810 to 1.217. From the visualization comparisons, we observe that it reduces lots of noisy surfaces.

\noindent\textbf{Pseudo-view Feature Consistency.} Although stereo geometry-texture alignment provides geometric guidance, it only regularizes Gaussians in the sparse training views and lacks the multi-view consistency, therefore, there are still changes to encounter overfitting issues. Adding the pseudo-view feature consistency further reduces the distances from 1.217 to 0.969. From the visualization results, the erroneous reconstructions are reduced, and more surface details are reconstructed.

\noindent\textbf{Train-view Feature Alignment.} With sparse views, the pseudo-view rendering may introduce additional noise due to the inherent lack of observation. Therefore, we further add feature alignment between training views to improve the robustness of the model to pseudo-view noise and further improve the surface details from the distance of 0.969 to 0.886.

\section{Conclusion}

This paper proposes \net{}, which enables accurate surface reconstruction from sparse views while simultaneously enhancing novel view synthesis. \net{} leverages Stereo Geometry-Texture Alignment to derive depth priors from rendered stereo pairs. We also introduce Pseudo-Feature Enhanced Geometry Consistency, which improves pseudo-view consistency through multi-view feature alignment. Extensive experiments on DTU, BlendedMVS, and Mip-NeRF360 demonstrate that \net{} achieves state-of-the-art results in sparse-view surface reconstruction.

\section{Acknowledgments}
This work was supported by the National Natural Science Foundation of China (Grant No.62276016, 62372029, 62306247), the Natural Science Foundation of Sichuan Province (2024NSFSC1474, 2024ZHCG0166).

\bibliography{aaai2026}

\clearpage

\appendix
\setcounter{secnumdepth}{2} 
 \begin{strip}
  \centering
  {\LARGE \textbf{Supplementary Material for \textit{SparseSurf: Sparse-View 3D Gaussian Splatting for Surface Reconstruction}}}
  \vspace{1em} 
\end{strip}
\section*{Overview}

We organize the material as follows. Sec.~\ref{supp_sec: experimental details} shows more details of the experimental settings. We provide more experiments and discussions in Sec.~\ref{supp_sec: additional ablations}. We present more comparisons on Mip-NeRF360 in Sec.~\ref{supp_sec: additional mipnerf360}. And we show more visualization results of \net{} in Sec.~\ref{supp_sec: additional visualization}. We provide the discussion of limitations in Sec.~\ref{supp_sec: limitation} and society impacts in Sec.~\ref{supp_sec: society impacts}.

\section{Details of Experimental Setting} \label{supp_sec: experimental details}

\subsection{Datasets}

\noindent\textbf{DTU.} 
The DTU dataset \cite{jensen2014large} comprises 124 object-centric scenes captured by a set of fixed cameras. For surface reconstruction, we follow the previous work \cite{huang2024neusurf, huang2025fatesgs} by selecting 15 scans with IDs 24, 37, 40, 55, 63, 65, 69, 83, 97, 105, 106, 110, 114, 118 and 122, which we split into two camera overlap settings, with views 22, 25 and 28 used as inputs for the little overlap (PixelNeRF) setting and views 23, 24 and 33 used as inputs for the large overlap (SparseNeuS) setting. The images are downsampled to half resolution.

For novel view synthesis we adopt the evaluation split of prior work \cite{li2024dngaussian, zhu2023fsgs, zhang2024cor, han2024binocular, zheng2025nexusgs} by choosing 15 scans with IDs 8, 21, 30, 31, 34, 38, 40, 41, 45, 55, 63, 82, 103, 110 and 114, in each of which images 25, 22 and 28 serve as the input views for rendering unseen viewpoints. The test set comprises images with IDs 1, 2, 9, 10, 11, 12, 14, 15, 23, 24, 26, 27, 29, 30, 31, 32, 33, 34, 35, 41, 42, 43, 45, 46 and 47 for quantitative evaluation. The images are downsampled to quarter resolution.

\noindent\textbf{BlendedMVS.} On the BlendedMVS dataset \cite{blendedmvs}, we employ the sparse-view subset processed by Neusurf \cite{huang2024neusurf}, which comprises eight challenging scenes rendered at 768 × 576 resolution, with three input views selected per scene.

\noindent\textbf{Mip-NeRF360.} Following prior sparse‐view methods \cite{niemeyer2022regnerf, zhu2023fsgs, zhang2024cor, zheng2025nexusgs}, we split Mip-NeRF360 \cite{mipnerf360} by selecting every 8-th image as a test view and evenly sampling 24 input views from the remaining images. The images are downsampled to quarter resolution, aligning with previous studies \cite{zhu2023fsgs,zhang2024cor}.

\subsection{Baselines}

We compare our model with state-of-the-art methods from five categories: (1) MVS methods: COLMAP \cite{colmap} and TransMVSNet \cite{ding2022transmvsnet}; (2) Generalizable sparse-view neural implicit reconstruction methods: SparseNeuS \cite{sparseneus}, VolRecon \cite{volrecon}, ReTR \cite{liang2024retr}, C2F2NeuS \cite{xu2023c2f2neus}, GenS \cite{peng2023gens} and UFORecon \cite{na2024uforecon}; (3) Per-scene optimization neural implicit methods: NeuS \cite{neus}, VolSDF \cite{volsdf}, MonoSDF \cite{monosdf} and NeuSurf \cite{huang2024neusurf}; and (4) Per-scene optimization Gaussian Splatting based methods: 3DGS \cite{3dgs}, DNGaussian\cite{li2024dngaussian}, CoR-GS\cite{zhang2024cor}, Gaussian Surfels \cite{gaussiansurfels}, 2DGS \cite{2dgs}, PGSR\cite{pgsr}, FatesGS \cite{huang2025fatesgs}, and concurrent work Sparse2DGS \cite{wu2025sparse2dgs}. 
For a fair comparison, we directly report the best quantitative results of most methods in corresponding published papers for comparisons. The results of 3DGS, Gaussian Surfels, and 2DGS are reported from FatesGS \cite{huang2025fatesgs}. 

\begin{figure*}[!t]
\centering
  \includegraphics[width=1.0\linewidth]{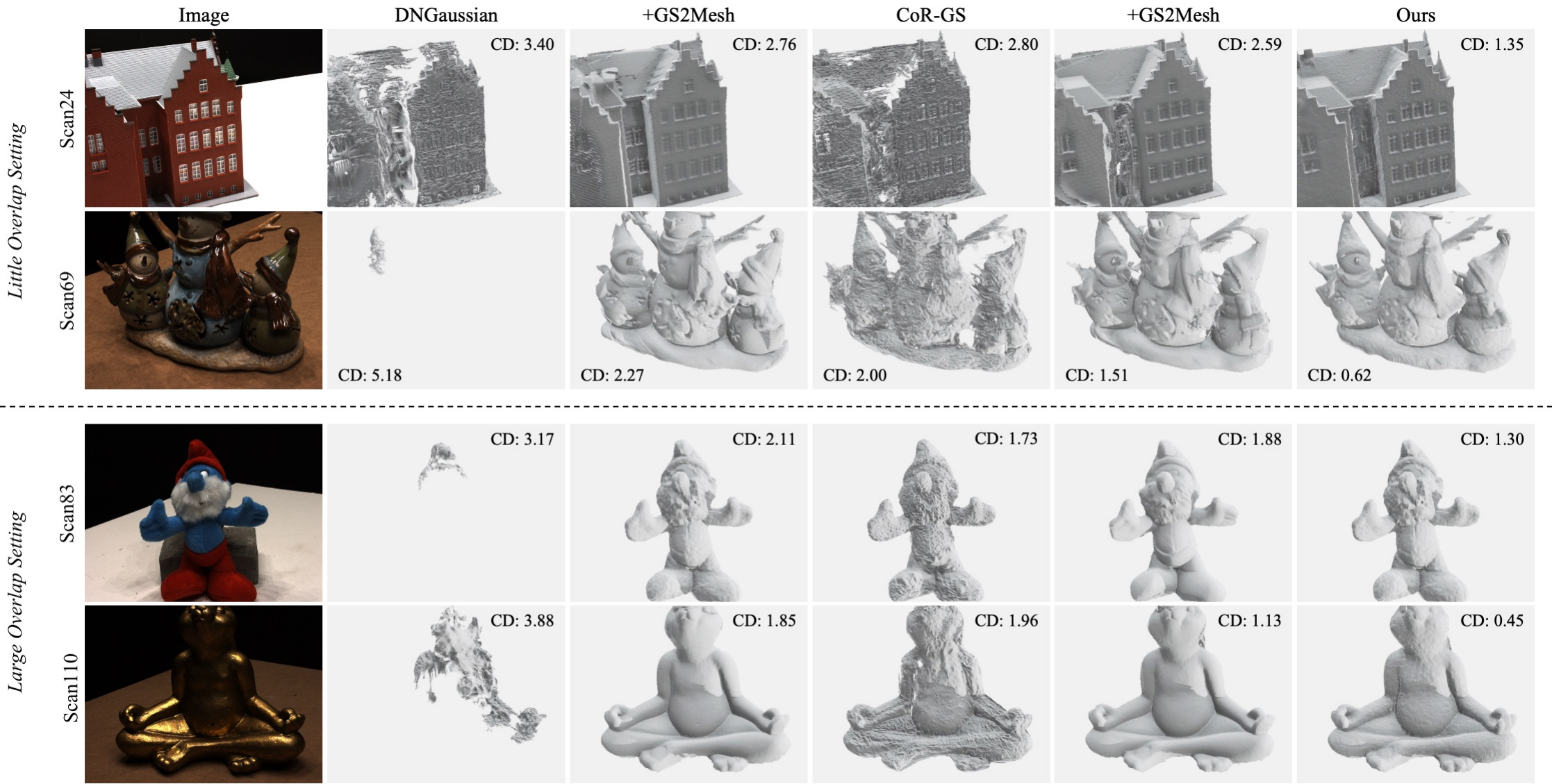}
    \caption{Qualitative comparison with GS2Mesh on DTU dataset. }
     \vspace{0cm} 
    \label{fig:gs2mesh}
\end{figure*}

\subsection{Implementations}

\noindent\textbf{Loss Functions. } 
Our overall loss function comprises the following terms: a ground‐truth color supervision $\mathcal{L}_{c}$, a depth–normal consistency loss $\mathcal{L}_{dn}$ between rendered normals and depth normals, a scale loss $\mathcal{L}_{s}$ to flatten Gaussians, and the proposed stereo alignment loss $\mathcal{L}_{stereo}$ to jointly align geometry and appearance, feature‐distillation loss $\mathcal{L}_{f}$, pseudo‐view feature consistency loss $\mathcal{L}_{pseudo}$, and train‐view feature alignment loss $\mathcal{L}_{train}$. Therefore, the total loss is:

\begin{equation}
\mathcal{L} = \mathcal{L}_{c} + \mathcal{L}_{stereo} + \lambda_{1} \mathcal{L}_{f} 
+\lambda_{2} \mathcal{L}_{pseudo} +\lambda_{3} \mathcal{L}_{train} +\lambda_{4} \mathcal{L}_{s}+\lambda_{5} \mathcal{L}_{dn}
\end{equation}

\noindent\textbf{Hyperparameters.} We train \net{} with 7000 iterations in total on a single NVIDIA RTX 3090 GPU. The $\mathcal{L}_{c}$, $\mathcal{L}_{s}$, and $\mathcal{L}_{f}$ are applied from the beginning of the training. The $\mathcal{L}_{stereo}$ is added from 500 iterations.
The $\mathcal{L}_{pseudo}$ and $\mathcal{L}_{dn}$ are applied from 3000 iterations. 
For the total loss functions, the loss weights $\lambda_{1}$, $\lambda_{2}$, $\lambda_{3}$, $\lambda_{4}$ and $\lambda_{5}$ are set to 1.5, 0.15, 1.5, 100, and 0.05, respectively. In the stereo loss $\mathcal{L}_{stereo}$, $\lambda_d$, $\lambda_n$, $\lambda_{nd}$, and  $\lambda_s$  are set to 0.05.

\noindent\textbf{Mesh Extraction.} Similar to previous surface reconstruction methods \cite{2dgs, pgsr, huang2025fatesgs}, we render depth from Gaussians and use truncated signed distance function (TSDF) fusion to extract the mesh. During TSDF fusion, the voxel size is set to 0.002.

\section{Additional Experiments and Analysis} \label{supp_sec: additional ablations}

\subsection{Ablation on Different Pre-trained Stereo-Matching Models.}

We evaluate our model with different pre-trained Stereo-Matching models, including Stereo Anywhere \cite{stereoanywhere} and Foundation Stereo \cite{wen2025stereo}. The evaluations are conducted on both little-overlap and large-overlap 3-view DTU settings. We provide the results in Table \ref{tab:pretained_stereo}. Compared to using Foundation Stereo, using Stereo Anywhere to provide stereo depth gets slightly worse results. However, the performances are still better than the state-of-the-art method FatesGS \cite{huang2025fatesgs} which uses the monocular depth prior. The comparison demonstrates the effectiveness of using stereo depth with accurate metric information and the robustness of our method to different stereo network architectures.

\begin{table} [h]

    \centering
    \resizebox{1.0\linewidth}{!}{
    \begin{tabular}{llcccc}
    \toprule
        Setting & Methods & Pretrained Model  & Accuracy$\downarrow$ & Completion$\downarrow$ & Average$\downarrow$  \\
    \midrule
     \multirow{3}{*}{Little-overlap}
     & FatesGS (ref) & Marigold  & - & - & 1.37 \\
     & \multirow{2}{*}{SparseSurf} & Stereo Anywhere & 0.763&1.445 & 1.104   \\
     & & Foundation Stereo & \textbf{0.724}&\textbf{1.376} &\textbf{1.050} \\
     \midrule
     \multirow{3}{*}{Large-overlap}
     & FatesGS (ref) & Marigold  & - & - & 0.92 \\
     & \multirow{2}{*}{SparseSurf} & Stereo Anywhere &0.545 &1.262 & 0.904   \\
     & & Foundation Stereo &\textbf{0.533} &\textbf{1.239} &\textbf{0.886} \\
    \bottomrule
    \end{tabular}}
    \caption{Impact of Pretrained Stereo Matching models on DTU dataset.}
    \label{tab:pretained_stereo}
\end{table}

\subsection{Ablation on Different Stereo Baseline.}
We conduct an ablation study by varying the horizontal baseline $b$ to 3\%, 7\%, and 10\% of the scene radius. As reported in Table \ref{tab:stereo_baseline}, our model achieves its optimal performance at $b=$3\% and it remains consistently strong at larger baselines, demonstrating robustness to baseline selection.

\begin{table} [h]
    \centering
    \resizebox{1.0\linewidth}{!}{
    \begin{tabular}{lccccc}
    \toprule
        Setting & Baseline &  Accuracy$\downarrow$ & Completion$\downarrow$ & Average$\downarrow$  \\
    \midrule
     \multirow{3}{*}{Little-overlap}
     & 3\%  & \textbf{0.724}&\textbf{1.376} &\textbf{1.050} \\
     & 7\% & 0.742 & 1.383 & 1.063   \\
     & 10\% & \textbf{0.724} & 1.454 & 1.089\\
     \midrule
     \multirow{3}{*}{Large-overlap}
     & 3\%  & {0.533} &\textbf{1.239} &\textbf{0.886} \\
     & 7\%  & {0.463} & 1.322 & 0.893  \\
     & 10\% & \textbf{0.424} & 1.391& 0.908 \\
    \bottomrule
    \end{tabular}}
    \caption{Impact of stereo horizontal baseline on DTU dataset.}
    \label{tab:stereo_baseline}
\end{table}

\begin{table*}[t]
    \centering
    \resizebox{1.0\linewidth}{!}{
        \begin{tabular}{l|*{15}{c}|c}
            \toprule
            Scan ID & 24 & 37 & 40 & 55 & 63 & 65 & 69 & 83 & 97 & 105 & 106 & 110 & 114 & 118 & 122 & Mean \\
            \midrule
            \midrule
            \multicolumn{17}{c}{\emph{Little-overlap (PixelNeRF) Setting} } \\
            \midrule
             DNGaussian(Random)& 3.40 & 5.58 & 3.28 &4.62&3.18&2.70&6.19&4.84&9.04&3.29&8.10&11.28& 3.90&4.86&2.69&5.13\\
             \quad+GS2Mesh&2.76&6.73&2.27&2.60&3.29&1.80&4.42&4.17&6.32&2.69&4.18&6.17&1.65&4.69&3.26&3.80\\
             DNGaussian&6.97&8.02&5.06&3.97&5.66&2.55&5.18&3.54&10.22&4.62&5.16&4.71&3.89&6.29&4.52&5.36\\
             \quad+GS2Mesh&6.74&8.11&4.20&4.82&5.39&2.20&2.27&3.01&9.59&2.47&4.00&2.98&3.57&5.86&4.15&4.62\\
             \cmidrule(l{0.7em}r{0.7em}){1-17}
            CoR-GS&2.80 &3.06 &2.42 &1.88&2.37&2.76&2.00&5.66&2.51&1.92&2.54&2.20&1.36&2.28&1.74&2.50\\
            \quad+GS2Mesh &2.59&3.26&2.76&1.08&1.94&2.19&1.51&2.22&2.60&1.55&2.18&1.93&0.63&1.68&1.46&1.97\\
            \cmidrule(l{0.7em}r{0.7em}){1-17}
            \textbf{Ours} &  \textbf{1.35} &  \textbf{2.10}&  \textbf{1.84} &  \textbf{0.66} & \textbf{0.79} &  \textbf{1.62} &  \textbf{0.62} &  \textbf{1.13} & \textbf{1.06} &  \textbf{0.76} &  \textbf{0.90} &  \textbf{0.87} &  \textbf{0.39} &  \textbf{0.70} & \textbf{0.96} &  \textbf{1.05} \\
        
            \midrule
            \midrule
            \multicolumn{17}{c}{\emph{Large-overlap (SparseNeuS) Setting}} \\
            \midrule
            DNGaussian(Random) & 3.25&3.49&5.10&1.61&6.45&2.94&5.60&5.30&5.45&2.54&4.05&3.88&1.91&7.53&6.64&4.38 \\
            \quad+GS2Mesh&2.32&4.26&5.25&1.68&6.38&2.82&1.45&6.83&2.04&1.54&1.73&1.85&0.82&1.48&1.46&2.79\\
            DNGaussian&3.88&5.46&3.60&1.62&4.92&1.92&5.17&3.17&5.29&3.01&2.97&4.66&1.73&3.62&2.21&3.55\\
            \quad+GS2Mesh&1.69&3.71&1.80&0.94&2.61&1.33&1.23&2.11&3.24&1.25&1.56&1.89&0.63&1.19&1.17&1.76\\
            \cmidrule(l{0.7em}r{0.7em}){1-17}
            CoR-GS&1.38&2.81&1.78&0.90&1.94&1.92&1.58&1.73&2.67&1.07&1.74&1.96&0.68&1.52&1.26&1.66\\
            \quad+GS2Mesh&1.13&2.58&1.56&0.84&1.62&1.40&1.00&1.88&1.89&0.93&1.03&1.13&0.66&1.04&1.22&1.33\\
            \cmidrule(l{0.7em}r{0.7em}){1-17}
            \textbf{Ours} & \textbf{0.72} & \textbf{1.61}& \textbf{1.17} & \textbf{0.72} & \textbf{1.11} & \textbf{1.13} & \textbf{0.57} & \textbf{1.30} & \textbf{1.20} & \textbf{0.67} & \textbf{0.70} & \textbf{0.45} & \textbf{0.36} & \textbf{0.70} & \textbf{0.87} & \textbf{0.89} \\
            \bottomrule
        \end{tabular} 
        }
    \caption{Comparisons with GS2Mesh \cite{wolf2024gs2mesh} on DTU dataset with two overlap settings. We test the method with Foundation Stereo as the stereo matching model. Best results are in bold. }
    \label{tab:gs2mesh}
\end{table*}

\begin{table*}[!ht]
    \centering
    \resizebox{1.0\linewidth}{!}{
        \begin{tabular}{l|*{15}{c}|c}
            \toprule
            Scan ID & 24 & 37 & 40 & 55 & 63 & 65 & 69 & 83 & 97 & 105 & 106 & 110 & 114 & 118 & 122 & Mean \\
            \midrule
            \midrule
            \multicolumn{17}{c}{\emph{3 Views (22, 25, 28)} } \\
            \midrule
            2DGS & 3.25 & 3.64 & 3.52 & 1.42 & 2.04 & 2.52 & 1.99 & 2.69 & 2.55 & 1.79 & 2.92 & 4.50 & 0.73 & 2.38 & 1.79 & 2.52 \\
            PGSR&5.83 &4.59 &4.52 &3.36&4.25&3.75&2.81&5.92&4.60&4.27&3.61&6.09&1.02&2.55&2.32&3.97\\
            FatesGS&  \textbf{1.32} &  2.85 &  2.71 &  0.80 &  1.44 & 2.08 &  1.11 &  1.19 &  1.33 &\textbf{0.76} &  1.49 &  \textbf{0.85} &  0.47 & 1.05 & 1.06 &1.37 \\
            \textbf{Ours} & 1.35 &  \textbf{2.10}&  \textbf{1.84} &  \textbf{0.66} & \textbf{0.79} &  \textbf{1.62} &  \textbf{0.62} &  \textbf{1.13} & \textbf{1.06} &  \textbf{0.76} &  \textbf{0.90} &  0.87 &  \textbf{0.39} &  \textbf{0.70} & \textbf{0.96} &  \textbf{1.05} \\
        
            \midrule
            \midrule
            \multicolumn{17}{c}{\emph{6 Views (22, 25, 28, 40, 44, 48)}} \\
            \midrule
            2DGS  &1.11&2.25&1.65&0.64&1.14&1.25&1.03&1.18&1.47&0.79&0.97&2.06&0.54&0.96&0.76&1.19\\
            PGSR&0.81&3.00&1.83&0.55&1.14&\textbf{0.90}&0.73&1.09&1.51&0.87&0.80&1.07&0.65&0.54&0.51&1.07\\
            FatesGS &0.58&2.41&1.18&\textbf{0.49}&1.02&1.37&0.79&1.22&1.34&0.77&0.71&0.64&0.35&0.44&0.53&0.92\\
            \textbf{Ours} & \textbf{0.48} & \textbf{1.92}& \textbf{0.70} & 0.51 & \textbf{0.67} & {0.91} & \textbf{0.45} & \textbf{1.02} & \textbf{0.84} & \textbf{0.66} & \textbf{0.58} & \textbf{0.52} & \textbf{0.28} & \textbf{0.41} & \textbf{0.48} & \textbf{0.70} \\
            \midrule
            \midrule

            \multicolumn{17}{c}{\emph{9 Views (0, 8, 13, 22, 25, 28, 40, 44, 48)}} \\
            \midrule
            2DGS &0.65&1.58&1.25&0.58&0.93&0.88&0.85&1.09&1.49&0.60&0.77&1.26&0.49&0.74&0.67&0.92\\
            PGSR&0.52&1.38&0.98&0.53&0.92&\textbf{0.64}&0.58&\textbf{1.04}&1.21&\textbf{0.59}&0.62&0.93&0.38&0.53&\textbf{0.44}&0.75\\
            FatesGS &0.49&1.21&0.71&\textbf{0.42}&0.91&1.15&0.64&1.27&1.31&0.67&0.74&0.91&0.31&\textbf{0.42}&0.50&0.78\\
            \textbf{Ours} & \textbf{0.40} & \textbf{1.13}& \textbf{0.62} & 0.45 & \textbf{0.58} & {0.81} & \textbf{0.47} & {1.05} & \textbf{0.94} & \textbf{0.59} & \textbf{0.49} & \textbf{0.71} & \textbf{0.29} & 0.47 & {0.45} & \textbf{0.63} \\
            \bottomrule
        \end{tabular} 
        }
    \caption{Comparisons on Different Input Views. Best results are in bold. }
    \label{tab:more_views}
\end{table*}

\subsection{Comparison with GS2Mesh.}
GS2Mesh \cite{wolf2024gs2mesh} is a dense-view surface reconstruction method that employs a pretrained stereo‐matching model to extract meshes rather than relying on depth maps rendered from the vanilla 3DGS representation, enabling accurate and complete geometry reconstruction from dense view inputs. However, when only sparse views are available, the 3DGS renderer tends to overfit those limited views, making it challenging to generate the high‐fidelity stereo pairs required for surface reconstruction. While our \net{} renders stereo-view images and feeds them into a pretrained stereo‐matching model to obtain geometric priors. As training progresses, the quality of stereo renderings improves, enhancing the accuracy of the stereo priors.

To evaluate GS2Mesh under sparse‐view setting, we integrate its mesh extraction via stereo matching into two sparse‐view 3DGS methods, DNGaussian \cite{li2024dngaussian} and CoR-GS \cite{zhang2024cor}. For a fair comparison, we use the state-of-the-art stereo-matching method Foundation Stereo \cite{wen2025stereo} to estimate depth maps. Table ~\ref{tab:gs2mesh} shows the comparisons on DTU dataset with the little overlap setting and the large overlap setting. Applying GS2Mesh significantly reduces the overall CD of both DNGaussian and CoR-GS, but our method still outperforms them due to the proposed Stereo Geometry-Texture Alignment. As illustrated in Figure ~\ref{fig:gs2mesh}, DNGaussian alone often fails to extract a coherent mesh in challenging scenes, but when augmented with GS2Mesh, it recovers a more complete surface. However, deriving its geometry directly from the stereo network’s depth maps, GS2Mesh leads to over-smoothing and blurring. In scan 83, CoR-GS combined with GS2Mesh produces a much smoother mesh and a worse CD. While our meshes achieve both more surface coverage and sharper geometric details.

\subsection{Comparison on Different Input Views.}

Table ~\ref{tab:more_views} presents a quantitative comparison across different numbers of input views on DTU, evaluating the dense‐view methods 2DGS \cite{2dgs} and PGSR \cite{pgsr} alongside the sparse‐view method FatesGS \cite{huang2025fatesgs}. Our method achieves the best average CD when using 3, 6, and 9 input views. 

\begin{figure*}[!ht]
\centering
  \includegraphics[width=0.95\linewidth]{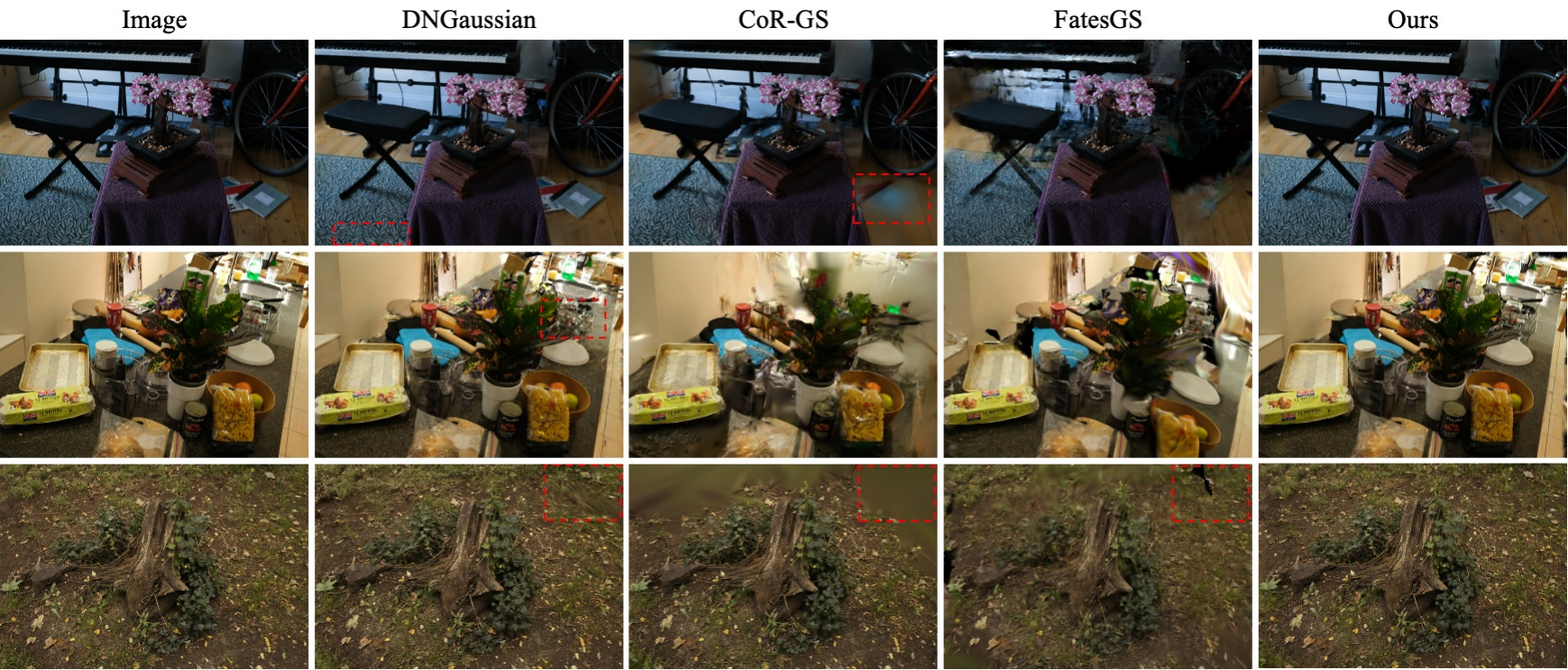}
    \caption{Qualitative rendering comparison on Mip-NeRF360 with 24 input views.}
     \vspace{0cm} 
    \label{fig:compare_m360_rend}
\end{figure*}

\subsection{Performance and Efficiency.}
We compare surface reconstruction performance and training time with previous state-of-the-art methods on the DTU large-overlap setting. As shown in Table \ref{tab:efficiency}, our method achieves the best performance on DTU dataset, with slightly slower runtime.
\begin{table}[!ht]
    \centering
    \begin{tabular}{l|cc}
    \toprule
        Method & Chamfer Distance & Training Time  \\
        \midrule
        NeuSurf & 0.99 & 14 hours \\
        FatesGS & 0.92 & 14 mins \\
        Sparse2DGS & 1.13 & \textbf{10 mins} \\
        \textbf{Ours} & \textbf{0.89} & {16 mins}  \\
        
    \bottomrule
    \end{tabular}
    \caption{Performance and efficiency. }
    \label{tab:efficiency}
\end{table}

\section{Additional Comparisons on Mip-NeRF360} \label{supp_sec: additional mipnerf360}

\subsection{Quantitative Comparison of Novel View Synthesis}
We conduct the sparse-view novel view synthesis experiments on Mip-NeRF360.Following CoR-GS\cite{zhang2024cor}, we use 24 views of each scene as input. Table \ref{tab:mip_nvs} shows that our method achieves the highest SSIM on Mip-NeRF360, indicating superior structural coherence and edge fidelity.

\begin{table}[!ht]
    \centering
    \tabcolsep=0.4cm
    \begin{adjustbox}{width=0.5\textwidth,center}
            \begin{tabular}{@{}l|cccc}
            
                \toprule
                \multirow{2}{*}{Method}     &  \multicolumn{4}{c}{Mip-NeRF360 (24 Views)}    \\
                \cmidrule(lr){2-5}
                & PSNR$\uparrow$ & SSIM$\uparrow$ & LPIPS$\downarrow$ & AVGE$\downarrow$  \\
                \midrule
                Mip-NeRF  & 21.23 & 0.613 & 0.351 & 0.118\\
                DietNeRF  & 20.21 & 0.557 & 0.387 &0.135 \\
                RegNeRF   & 22.19 & 0.643 & 0.335 & 0.107\\
                FreeNeRF  & 22.78 & 0.689 & 0.323 & 0.098\\
                SparseNeRF  & 22.85 & 0.693 & 0.315 & 0.097\\ 
                DS-NeRF   & 14.58 & 0.311 & 0.692 & 0.271\\              
                ViP-NeRF  & 14.78 & 0.300 & 0.774 & 0.278 \\ 
                SimpleNeRF  &10.82 & 0.142 & 0.880 &  0.407 \\ 
                \midrule
                3DGS    & 0.672 & 0.248 &0.099 \\
                DNGaussian   & 18.06 & 0.423 & 0.584 &0.191\\           
                CoR-GS  & \cellcolor{orange}23.55 & \cellcolor{yellow}0.727 &\cellcolor{yellow} 0.226 & \cellcolor{orange}0.080 \\ 
                NexusGS &   \cellcolor{red}23.86 & \cellcolor{orange}0.753 & \cellcolor{red} 0.206 & \cellcolor{red}0.075\\
                \midrule
                FatesGS  &  17.17&0.512&0.462&0.197\\
                \textbf{Ours}   &\cellcolor{yellow}23.01&\cellcolor{red}0.760&\cellcolor{orange}0.210&\cellcolor{yellow}0.086\\
                \bottomrule
            \end{tabular}
            \end{adjustbox}
        \caption{Quantitative evaluations of render quality on Mip-NeRF360 datasets with sparse-view NVS setting. }
        \label{tab:mip_nvs}
\end{table}

\begin{figure*}[!ht]
\centering
  \includegraphics[width=0.95\linewidth]{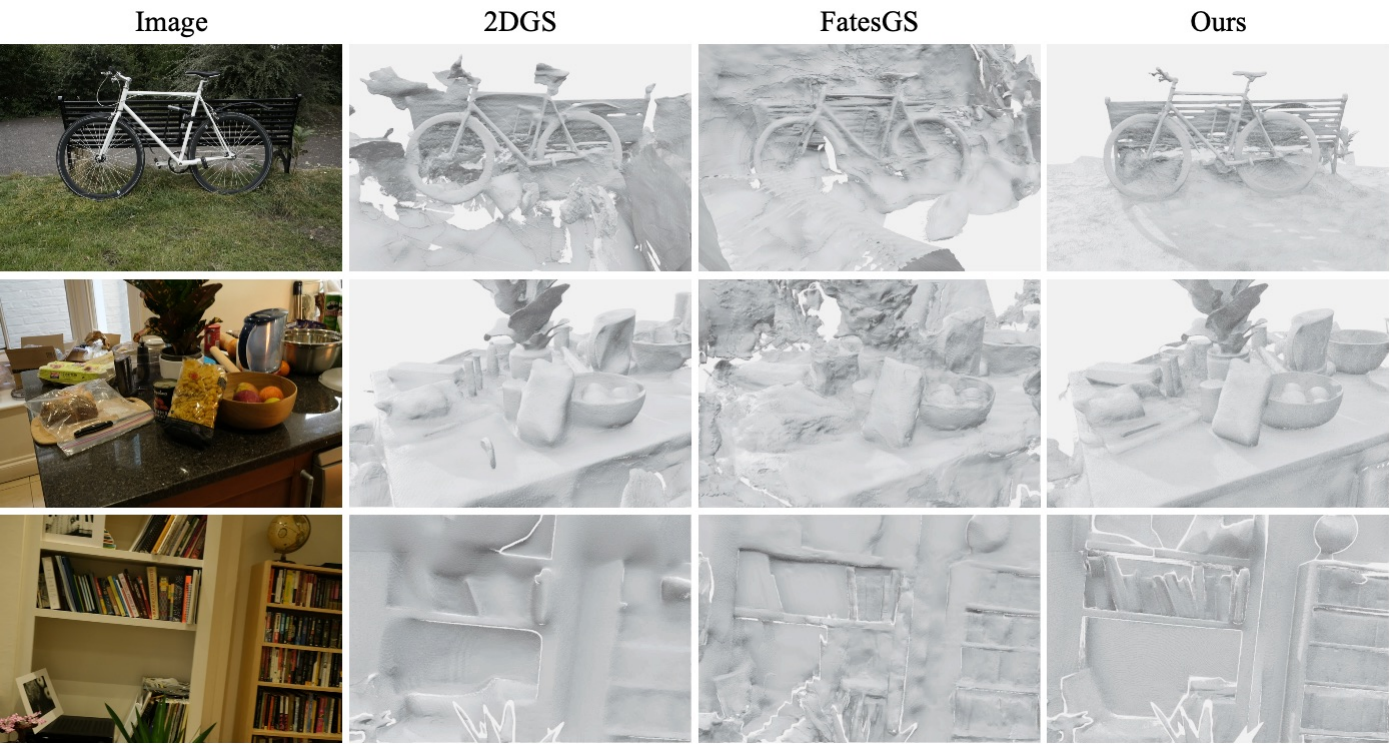}
    \caption{Qualitative mesh comparison on Mip-NeRF360 with 24 input views.}
     \vspace{0cm} 
    \label{fig:compare_m360_mesh}
\end{figure*}

\subsection{Qualitative Comparison of Novel View Synthesis}
Figure \ref{fig:compare_m360_rend} illustrates the visual comparison of rendering quality on Mip-NeRF360. Our method renders sharp novel-view images that preserve both the overall scene appearance and fine-grained textures while exhibiting fewer artifacts.

\begin{figure*}[ht!]
\centering
  \includegraphics[width=1.0\linewidth]{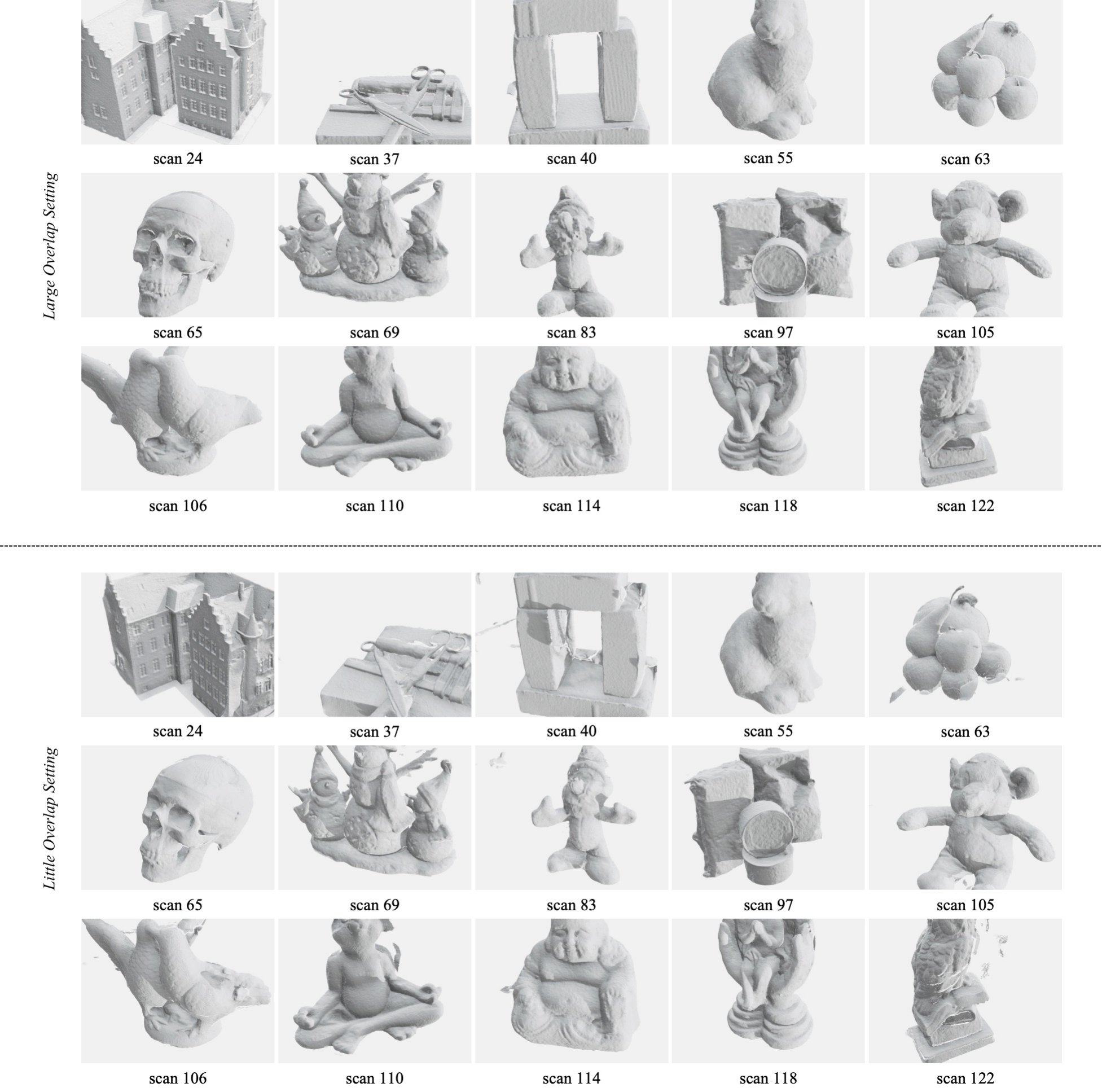}
    \caption{Visualized results of surface reconstruction on DTU dataset.}
     \vspace{0cm} 
    \label{fig:supp_dtu}
\end{figure*}

\subsection{Qualitative Mesh Comparison}

Figure \ref{fig:compare_m360_mesh} shows the visual comparison of surface reconstruction on Mip-NeRF360. Our meshes achieve richer geometric detail than 2DGS and FatesGS, and avoid the background adhesion.

\section{Additional Visualization Results} \label{supp_sec: additional visualization}

We present our visualization results on all 15 tested scenes from the DTU dataset under large overlap
setting and little overlap setting in Figure ~\ref{fig:supp_dtu}.

\section{Limitation} \label{supp_sec: limitation}

Our method can reconstruct detailed and accurate surfaces from sparse input views while maintaining novel-view rendering quality. 
However, in the case of sparse-view, occlusion problems are inevitable, and surface reconstruction may fail in these regions.

\section{Society Impacts} \label{supp_sec: society impacts}

Our \net{}, a sparse‐view surface reconstruction method, lowers the barrier to high‐quality 3D reconstruction by reducing both the number of required input images and the need for specialized capture hardware, enabling cultural heritage institutions, and individual creators to reconstruct accurate 3D models of artifacts, architecture, and natural scenes at minimal cost. While making realistic scene reconstruction from limited views more accessible may raise privacy concerns if used improperly.

\end{document}